\documentclass{article}

\PassOptionsToPackage{dvipsnames,table}{xcolor}

\usepackage{iclr2027_conference,times}

\usepackage{latexsym}
\usepackage[T1]{fontenc}
\usepackage[utf8]{inputenc}
\usepackage{microtype}
\usepackage{inconsolata}

\usepackage{graphicx}
\usepackage{amsmath}
\usepackage{amssymb}
\usepackage{booktabs}
\usepackage{multirow}
\usepackage{adjustbox}
\usepackage{makecell}
\usepackage{siunitx}
\usepackage{xcolor}
\usepackage{array}
\usepackage{tabularx}
\usepackage{enumitem}
\usepackage{wrapfig}

\usepackage{hyperref}
\usepackage{url}
\usepackage{titletoc}

\newcommand{\deltaup}[1]{\textcolor{ForestGreen}{#1}}
\newcommand{\deltadown}[1]{\textcolor{BrickRed}{#1}}

\definecolor{citeblue}{rgb}{0.21,0.49,0.74}
\hypersetup{
    colorlinks=true,
    citecolor=citeblue,
    linkcolor=purple,
    urlcolor=purple
}

\newcommand{\ours}[1]{\textsc{MERIT}}

\definecolor{darkblue}{rgb}{0, 0, 0.5}

\definecolor{oursbg}{RGB}{230,231,250}
\newcommand{\ourcell}[1]{\cellcolor{oursbg}#1}

\newcommand{\traincell}[1]{%
  \multicolumn{1}{c}{\cellcolor{black!7}#1}%
}

\title{
Lost in Adaptation: Layer-Selective Recovery of Temporal Reasoning
\\in Video-Language Models
}

\author{
  \textbf{Zihang Fu}\textsuperscript{$\spadesuit$},
  \textbf{Haonan Wang}\textsuperscript{$\spadesuit$},
  \textbf{Jian Kang}\textsuperscript{$\clubsuit$},
  \textbf{Kenji Kawaguchi}\textsuperscript{$\spadesuit$},
  \textbf{Jiaying Wu}\textsuperscript{$\spadesuit$}\thanks{Corresponding author.}
  \\
  \textsuperscript{$\spadesuit$}National University of Singapore
  \quad
  \textsuperscript{$\clubsuit$}MBZUAI
  \\
  \texttt{zihangfu@u.nus.edu, jiayingwu@u.nus.edu}
}

\iclrfinalcopy 

\begin{document}

\maketitle

\vspace{-0.5em}
\begin{abstract}

Multimodal adaptation can weaken temporal reasoning (TR) in video-language models (VLMs) even when the visual evidence needed for an answer is correctly perceived. This weakened competence, however, remains partly recoverable from the model's paired pre-adaptation language checkpoint. We introduce \ours{}, a gradient-free framework that selectively restores this competence across the language backbone while protecting temporal perception (TP). It assigns each self-attention layer a VLM-dominant or LLM-dominant interpolation and uses the Covariance Matrix Adaptation Evolution Strategy (CMA-ES) to search the recipe space under a perception-constrained objective. Across three VLM families and five video benchmarks, \ours{} improves reasoning-aligned performance in all 15 model--benchmark evaluations while preserving TP across all three Video-MME settings. Recipes selected using 232 diagnostic examples transfer to four unseen benchmarks, with relative gains up to 27.8\%, and outperform full-backbone, uniform-attention, and alpha-tuned random-layer merging. Interventional masking shows that \ours{}-selected layers are disproportionately important for TR, while frame-level attribution reveals increased reliance on temporally distributed, causally relevant evidence. These results show that adaptation-induced reasoning degradation can be repaired post hoc through selective reuse of pre-adaptation competence, without learning new parameters or adding inference overhead.
\end{abstract}

\begin{figure}[t]
\vspace{-0.5em}
\centering
\includegraphics[width=0.95\columnwidth]{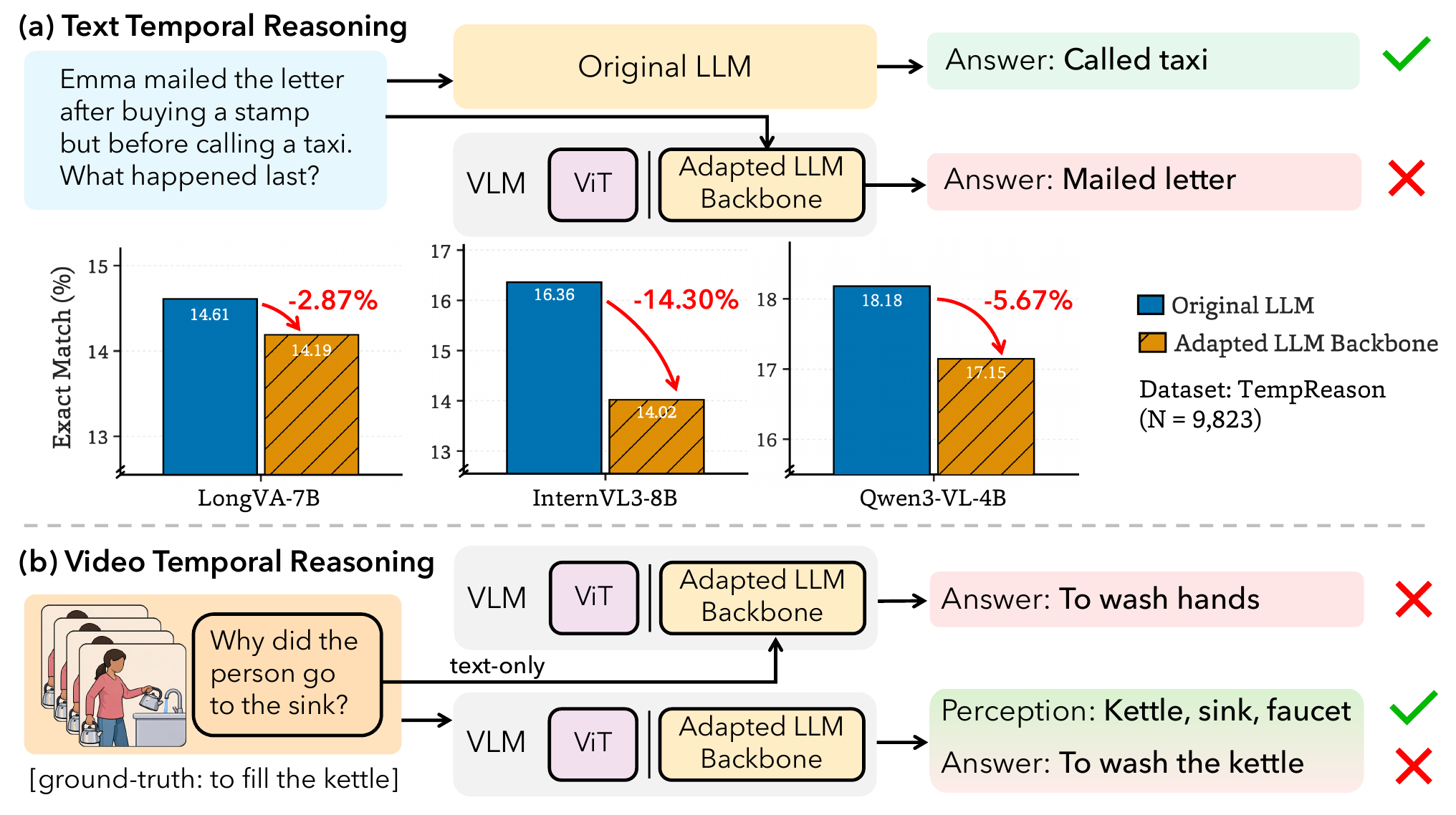}
\caption{\textbf{Multimodal adaptation can preserve what a VLM sees while weakening how its language backbone reasons over time.}
\textbf{(a)} The adapted language backbone underperforms its pre-adaptation LLM on text-only temporal reasoning (TR), directly isolating adaptation-induced degradation within the language component (\S\ref{app:tempreason}).
\textbf{(b)} In video TR, the VLM correctly perceives the relevant visual entities but fails to infer the temporal-causal structure required for the answer.}
\label{fig:intro_pic}
\vspace{-1.5em}
\end{figure}

\section{Introduction}

Multimodal adaptation equips pretrained language models with visual capabilities, but it can also weaken reasoning abilities acquired during language pretraining \citep{li2024multi}. We observe this regression in temporal reasoning (TR). As shown in Figure~\ref{fig:intro_pic}, the language backbone of an adapted video-language model (VLM) performs worse than its original LLM checkpoint on text-only TR, indicating that part of the degradation arises within the language component itself. In video settings, the same loss can appear as a gap between perception and reasoning: the model recognizes relevant objects and actions, yet fails to infer event order, state transitions, or causal dependencies. Such failures are particularly consequential for video temporal reasoning, which requires tracking temporal and causal structure \citep{chu2024timebench,wang2024tram} and integrating evidence across frames \citep{chen2024rextime,fang2024mmbench,nie2024slowfocus,cheng2025video}. This suggests that some video reasoning failures reflect capabilities weakened during multimodal adaptation, even when the relevant visual evidence remains available.

This observation suggests that temporal reasoning need not be learned anew: the original LLM checkpoint already provides a source of the competence weakened during multimodal adaptation. At the same time, the adapted VLM provides the perceptual grounding required for video understanding. The resulting challenge is to recover the former while preserving the latter. Fine-tuning can improve reasoning \citep{li2024multi}, but requires additional supervision and parameter updates. Model merging offers a direct post-hoc mechanism for recombining the complementary capabilities of the paired pre-adaptation and post-adaptation checkpoints. Prior work has explored evolutionary recipe search \citep{akiba2025evolutionary}, multimodal component integration \citep{sung2023empirical,zhu2025remedy}, and reasoning transfer through merging \citep{chen2025bring}. However, these approaches generally operate over entire backbones or predefined module groups, applying a similar compromise across the language stack. This granularity may be limiting, as reasoning behavior can be highly sensitive to particular layers or subnetworks \citep{sharma2024truth,shi2024hypothesis,ju2024large}. The key challenge is therefore to determine where the original reasoning capability should be restored while leaving adaptation-induced perceptual competence intact. This leads to our central question:

\vspace{-0.5em}
\begin{center}
\textbf{\textit{Where should pre-adaptation reasoning competence be restored to recover temporal reasoning while preserving multimodal perception?}}
\end{center}
\vspace{-0.5em}

We introduce \ours{} (\textit{\underline{M}erging for \underline{E}nhanced \underline{R}easoning \underline{I}n \underline{T}emporal Tasks}), a gradient-free framework for layer-selective capability recovery. Our key idea is to determine \emph{where} the adapted VLM should retain multimodal parameters and where it should draw more strongly on the original LLM. Specifically, \ours{} assigns each self-attention layer a VLM-dominant or LLM-dominant interpolation, producing a combinatorial space of merging recipes. Candidate recipes are evaluated with a perception-constrained objective that rewards TR improvement while penalizing temporal perception (TP) degradation. As this objective is non-differentiable with respect to discrete layer assignments, \ours{} uses the Covariance Matrix Adaptation Evolution Strategy (CMA-ES) \citep{hansen2001completely} to search the recipe space. All source parameters remain fixed throughout.

Across three VLM families and five video benchmarks, \ours{} consistently recovers temporal reasoning while preserving perceptual capability. Recipes selected using only 232 diagnostic examples improve reasoning-aligned performance across all 15 model--benchmark comparisons and transfer to four unseen benchmarks, with relative gains of up to 27.8\%. Evaluation on InternVL3-14B further shows that the same recovery pattern extends beyond the 4B--8B setting (\S~\ref{app:scalability}). Controlled comparisons show that the gains depend on the discovered layer configuration: \ours{} outperforms globally tuned full-backbone merging, uniform attention merging, and alpha-tuned random layer selection. Although LoRA and full-parameter tuning fit the diagnostic examples more strongly, their gains transfer less reliably, consistent with \ours{} recovering competence already encoded in the paired checkpoint. Interventional masking further shows that the selected layers are functionally important for temporal reasoning, while frame-level attribution indicates that the repaired models rely more strongly on temporally distributed and causally relevant evidence. These results establish adaptation-induced reasoning degradation as a tractable post-hoc repair problem and identify layer-selective merging as an effective mechanism for capability recovery.

\vspace{-0.5em}
\section{Related Work}
\vspace{-0.5em}

\textbf{Video Temporal Reasoning.}
Video temporal reasoning requires modeling event order, state changes, and causal relations across frames. Benchmarks including CLEVRER \citep{yi2019clevrer}, NExT-QA \citep{xiao2021next}, TempCompass \citep{liu2024tempcompass}, TOMATO \citep{shangguan2024tomato}, TVBench \citep{cores2024lost}, Vinoground \citep{zhang2024vinoground}, and LongVideoBench \citep{wu2024longvideobench} expose persistent weaknesses in temporal perception and reasoning. \citet{li2024temporal} further show that the language backbone can bottleneck video TR even when visual representations contain sufficient temporal information. \ours{} identifies a more specific source of this bottleneck: multimodal adaptation can weaken temporal competence already present in the pre-adaptation language backbone.

\textbf{Improving Reasoning in Video-Language Models.}
Training-based methods improve temporal reasoning through time-aware architectures and instruction tuning \citep{ren2024timechat,huang2024vtimellm}, self-training and distillation \citep{shi2025enhancing,qiu2025step,fei2024video}, or text-only temporal supervision \citep{li2024temporal}. Training-free methods modify inference through task decomposition \citep{min2024morevqa,liao2025divide}, constrained decoding \citep{qi2025improve}, or reasoning demonstrations \citep{wang2025harm}. \ours{} instead performs post-hoc parameter-level recovery using paired checkpoints, without gradient updates or additional inference procedures.

\textbf{Adaptive Model Merging.}
Model merging combines capabilities across checkpoints without conventional retraining. AdaMerging \citep{yang2024adamerging} learns task- or layer-wise merging coefficients, while evolutionary model merging \citep{akiba2025evolutionary} searches merging recipes through black-box optimization. \ours{} applies evolutionary search to a different objective: selecting layer-wise interventions between paired pre- and post-adaptation checkpoints while protecting temporal perception.

\textbf{Multimodal Model Merging and Capability Recovery.}
Prior multimodal merging work combines vision, language, and cross-modal components \citep{sung2023empirical}, allocates reusable multimodal modules \citep{zhu2025remedy}, or injects reasoning from specialized LLMs into VLMs \citep{chen2025bring}. Locate-then-Merge \citep{yu2025locatethenmerge} addresses multimodal catastrophic forgetting through neuron-level parameter fusion, while recent methods use adaptive cross-layer matching \citep{zhang2026chain} or weight-decomposed merging \citep{zou2026injecting}. \ours{} specifically recovers adaptation-induced temporal reasoning loss by searching for task-effective self-attention interventions in the VLM's own pre-adaptation backbone under an explicit perception-preservation objective.

\vspace{-0.5em}
\section{\ours{}: Layer-Selective Temporal Reasoning Recovery in VLMs}
\label{sec:approach}
\vspace{-0.5em}

\begin{figure*}[t]
\vspace{-0.5em}
\centering
\includegraphics[width=0.95\textwidth]{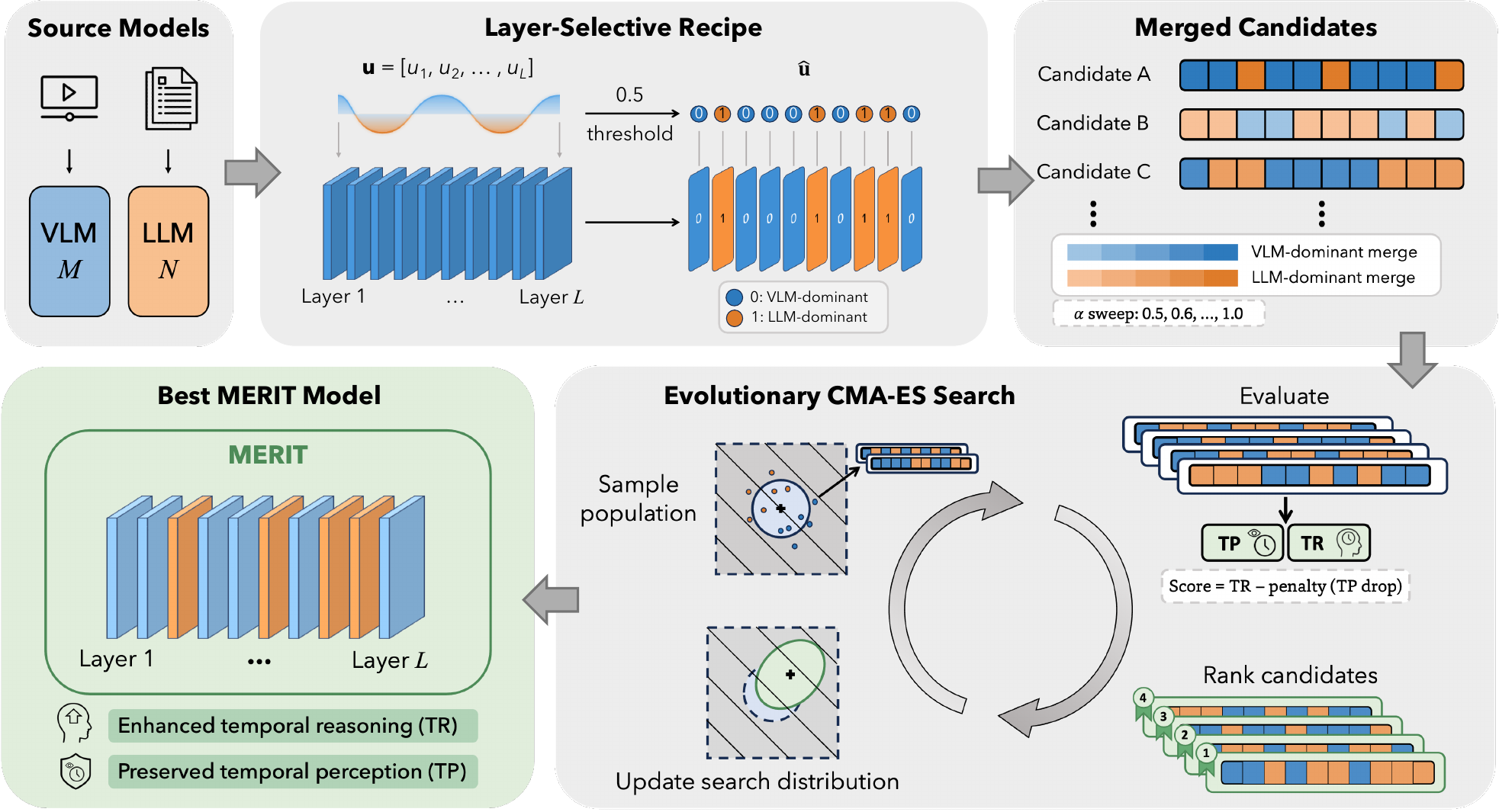}
\caption{\textbf{Overview of \ours{}.}
Given an adapted VLM and its paired pre-adaptation LLM, \ours{} searches for a layer-wise self-attention merging recipe in which each layer is VLM-dominant or LLM-dominant. CMA-ES optimizes this discrete structure under a perception-constrained objective that rewards TR recovery while penalizing TP degradation. The selected recipe is compiled into a single merged VLM with no additional inference components.}
\vspace{-1em}
\label{fig:main_pic}
\end{figure*}

\ours{} treats adaptation-induced reasoning degradation as a selective recovery problem. We assume access to an adapted VLM and its paired pre-adaptation language backbone. The two checkpoints contain complementary capabilities: the adapted model provides the visual grounding needed for video understanding, while the earlier language checkpoint retains reasoning competence weakened during multimodal adaptation. Globally moving toward either checkpoint can therefore trade reasoning against perception. \ours{} instead searches for a structured combination in which different self-attention layers draw to different degrees from the two checkpoints. The resulting recipe is selected to improve temporal reasoning (TR) while explicitly protecting temporal perception (TP), as illustrated in Figure~\ref{fig:main_pic}.

\vspace{-0.3em}
\subsection{Recovery Setting and Perception-Constrained Objective}
\label{subsec:objective}
\vspace{-0.3em}

Let $M$ denote an adapted VLM and $N$ its paired pre-adaptation text-only LLM, with a shared language architecture. This pairing isolates a single adaptation trajectory: $N$ provides the parameter state before multimodal adaptation, while $M$ contains the capabilities acquired afterward. Let $g$ denote a merging recipe that determines how the self-attention parameters of the two checkpoints are combined across the $L$ language layers. We parameterize $g$ explicitly in \S~\ref{subsec:layer_merging}. Let $\mathrm{Acc}_{t}(g)$ denote the accuracy of the resulting model on task $t\in\{\mathrm{TP},\mathrm{TR}\}$.

The desired repair is asymmetric. We seek to recover TR weakened during adaptation while avoiding loss of the TP competence already present in $M$. Therefore, we define perceptual degradation relative to the adapted VLM:
\begin{equation}
D_{\mathrm{TP}}(g)
=
\max\left(
0,\,
\mathrm{Acc}^{\mathrm{base}}_{\mathrm{TP}}
-
\mathrm{Acc}_{\mathrm{TP}}(g)
\right),
\label{eq:tp_degradation}
\end{equation}
where $\mathrm{Acc}^{\mathrm{base}}_{\mathrm{TP}}$ is the TP accuracy of $M$. The one-sided form penalizes only degradation relative to the adapted model; a recipe that also improves perception is not discouraged.

We then optimize
\vspace{-0.5em}
\begin{equation}
g^{\ast}=\arg\max_g \mathcal{F}(g),
\qquad
\mathcal{F}(g)=\mathrm{Acc}_{\mathrm{TR}}(g)-\lambda D_{\mathrm{TP}}(g),
\vspace{-0.2em}
\label{eq:tr_optimize}
\end{equation}
where $\lambda$ controls the strength of perceptual protection. This objective directly encodes the recovery criterion: a useful intervention should improve the weakened capability while preserving what multimodal adaptation added.

Importantly, TP and TR examples provide only black-box signals for selecting among candidate recipes. They do not update either checkpoint. Improvements must therefore arise by recombining competence already present in $M$ and $N$, instead of by fitting new parameters to the diagnostic set.

\vspace{-0.3em}
\subsection{Layer-Selective Self-Attention Merging}
\label{subsec:layer_merging}
\vspace{-0.3em}

A central question is where the pre-adaptation checkpoint should be reintroduced. A single global interpolation forces every layer to make the same trade-off between the two checkpoints, even though multimodal adaptation need not affect all regions of the language backbone equally. We therefore expose the layer structure itself as the decision space: some layers can remain close to the adapted VLM, while others can draw more strongly from the pre-adaptation LLM.

We intervene on self-attention because temporal reasoning requires selecting question-relevant evidence and integrating information across temporally separated visual tokens. Self-attention directly mediates these interactions, making it a natural target for selective recovery. Within self-attention, we use layers as the unit of selection to avoid the coarse granularity of global merging without incurring the large, architecture-specific search space of head-level selection. Layer-level recipes remain portable across model families while still allowing different regions of the backbone to draw on different capability sources. Notably, this formulation does not assume that TR is localized to individual layers; it asks which layer-level interventions provide effective recovery with limited perceptual interference.

We instantiate each recipe as
$g=(\hat{\mathbf{u}},\alpha)$,
where $\hat{\mathbf{u}}\in\{0,1\}^{L}$ specifies the layer-wise dominance pattern and $\alpha$ controls interpolation strength. To search over the discrete pattern, we parameterize it through a continuous gating vector $\mathbf{u}\in[0,1]^L$ and threshold each coordinate:
\begin{equation}
\hat{u}_{\ell}
=
\begin{cases}
1, & u_{\ell}\geq 0.5,\\
0, & u_{\ell}<0.5.
\end{cases}
\label{eq:threshold}
\end{equation}

Given $\hat{\mathbf{u}}$, the self-attention parameters at layer $\ell$ are constructed as
\begin{equation}
\theta_{\ell}
=
\begin{cases}
(1-\alpha)\theta^{M}_{\ell}
+
\alpha\theta^{N}_{\ell},
& \hat{u}_{\ell}=1,\\
\alpha\theta^{M}_{\ell}
+
(1-\alpha)\theta^{N}_{\ell},
& \hat{u}_{\ell}=0.
\end{cases}
\label{eq:directional_merge}
\end{equation}
The interpolation is defined for $\alpha\in[0,1]$. Equation~\ref{eq:directional_merge} has an exact complement symmetry:
$(\hat{\mathbf{u}},\alpha)$ and
$(\mathbf{1}-\hat{\mathbf{u}},1-\alpha)$
produce the same merged checkpoint. Thus, when layer assignments are allowed to vary, coefficients below $0.5$ introduce no additional checkpoints, and we restrict the search to $\alpha\in[0.5,1]$ without loss of expressivity. Within this range, $\hat{u}_{\ell}=1$ denotes an LLM-dominant layer and $\hat{u}_{\ell}=0$ a VLM-dominant layer. At $\alpha=0.5$, the two assignments coincide.

This parameterization separates two decisions that global merging conflates: \emph{where} the pre-adaptation checkpoint should dominate and \emph{how strongly} the two checkpoints should be mixed. It therefore allows \ours{} to preserve multimodal adaptation where it remains useful while restoring pre-adaptation parameters where they improve reasoning. We evaluate self-attention against MLP-only, attention-plus-MLP, and full-block interventions in \S~\ref{sec:design_validation}.

\vspace{-0.3em}
\subsection{Evolutionary Recipe Search}
\label{subsec:search}
\vspace{-0.3em}

Layer-selective recovery creates a combinatorial black-box optimization problem. For fixed $\alpha\neq0.5$, $L$ self-attention layers induce $2^L$ binary layer patterns; a 28-layer backbone already yields more than 268 million candidates. Moreover, layer decisions can interact, so the value of one assignment cannot be assessed independently of the others. Recipe quality is observable only after constructing the merged checkpoint and evaluating downstream TR and TP.

We optimize this objective using the Covariance Matrix Adaptation Evolution Strategy (CMA-ES) \citep{hansen2001completely}. For a fixed $\alpha$, CMA-ES maintains a distribution over continuous gating vectors $\mathbf{u}$, thresholds them into binary patterns $\hat{\mathbf{u}}$, instantiates the corresponding recipes, and evaluates their fitness with $\mathcal{F}$. Its adaptive covariance captures dependencies among layer decisions, enabling structured exploration of interacting interventions.

Exploiting the complement symmetry of Eq.~\ref{eq:directional_merge}, we search only one half of the interpolation range. We run an independent CMA-ES search for each
$\alpha\in\{1.0,\,0.9,\,0.8,\,0.7,\,0.6\}$.
At $\alpha=0.5$, the two branches of Eq.~\ref{eq:directional_merge} coincide, so the layer pattern has no effect and only one recipe is evaluated. We retain the recipe with the highest objective across all candidates. The outer grid controls mixing strength, while CMA-ES determines where each checkpoint should dominate. This separation also allows us to test whether gains arise from the discovered layer structure or merely from interpolation strength.

Once selected, the recipe is compiled directly into a single VLM checkpoint. \ours{} therefore incurs an offline search cost while leaving deployment unchanged: the recovered model has the same architecture and inference procedure as the original VLM, with no trainable adapters, ensembles, or additional decoding steps.

\vspace{-0.3em}
\section{Experiments}
\vspace{-0.3em}

We evaluate \ours{} through four questions about adaptation-induced capability recovery:

\begin{itemize}[leftmargin=*]
    \item \textbf{RQ1: Recovery and transfer (\S~\ref{sec:main_results}).}
    Is adaptation-induced TR loss recoverable while preserving TP, and does this recovery generalize beyond the diagnostic set?

    \item \textbf{RQ2: Determinants of recovery (\S~\ref{sec:design_validation}).}
    Does successful recovery depend on \emph{where} pre-adaptation parameters are restored, beyond interpolation strength or supervised fitting?

    \item \textbf{RQ3: Functional intervention (\S~\ref{sec:masking}).}
    Are the selected layers especially effective intervention points for temporal reasoning?

    \item \textbf{RQ4: Temporal grounding (\S~\ref{sec:attribution}).}
    Does recovery shift decisions toward temporally distributed, causally informative evidence?
\end{itemize}

\subsection{Experimental Setup}
\label{sec:exp_setup}

We evaluate \ours{} on LongVA-7B \citep{zhang2024long}, InternVL3-8B \citep{zhu2025internvl3}, and Qwen3-VL-4B-Instruct \citep{bai2025qwen3vltechnicalreport}. Each VLM is paired with the pre-adaptation language backbone specified by its original release. Exact checkpoints and architectures are provided in \S~\ref{sec:model_details}.

\textbf{Search and Evaluation Data.}
We use 55 Temporal Perception (TP) and 177 Temporal Reasoning (TR) examples from Video-MME \citep{fu2025video} solely for recipe selection and in-domain diagnostic evaluation. The remaining 2,468 Video-MME examples form the held-out \textsc{Others} subset. Once selected, each recipe is frozen and evaluated on four unseen benchmarks: LongVideoBench \citep{wu2024longvideobench}, LVBench \citep{wang2025lvbench}, MMBench-Video \citep{fang2024mmbench}, and Video-Holmes \citep{cheng2025video}. This separation tests whether a compact diagnostic signal can identify transferable recovery interventions. Dataset and metric details are provided in \S~\ref{sec:metric_details} and \S~\ref{sec:benchmark_details}.

\textbf{Compared Methods.}
We compare \ours{} with the unmodified VLM (\textsc{Base}); supervised \textsc{LoRA} and full-parameter fine-tuning (\textsc{FFT}) on the same 232 examples; \textsc{Full-Backbone} and \textsc{All-Attention} interpolation; and \textsc{Random-k} with either fixed or diagnostic-tuned interpolation strength. For both random-layer baselines, $k$ matches the number of layers selected by \ours{}, and we report mean and standard deviation over five runs. Full configurations are provided in \S~\ref{sec:baseline_details}.

\textbf{Search and Evaluation Protocol.}
We optimize layer-wise recipes with CMA-ES \citep{hansen2001completely} using \texttt{mergekit-evolve} \citep{goddard2024arcee}. Each fixed-$\alpha$ search evaluates at most 1,600 candidates and retains the recipe with the highest perception-constrained objective. All source parameters remain fixed, and evaluation uses deterministic decoding and official benchmark protocols. Further implementation details are provided in \S~\ref{sec:implementation_details}.

\begin{wraptable}{R}{0.6\textwidth}
    \centering
\vspace{-1.5em}
\caption{\textbf{\ours{} recovers temporal reasoning while preserving perception on Video-MME.}
Accuracy (\%) on the TP/TR diagnostic subsets and held-out \textsc{Others}. Shaded entries are in-sample; $\dagger$ denotes $\alpha$-tuned \textsc{Random-k}; parentheses show relative change over \textsc{Base}.}
\vspace{0.3em}
\small
\renewcommand{\arraystretch}{1.08}
\setlength{\tabcolsep}{5pt}
\begin{adjustbox}{max width=0.58\columnwidth}
\begin{tabular}{l l
                S[table-format=2.1]
                S[table-format=2.1]
                S[table-format=2.1]}
\toprule
\textbf{Model}
& \textbf{Variant}
& \multicolumn{2}{c}{\textbf{Diagnostic}}
& \multicolumn{1}{c}{\textbf{Held-out}} \\
\cmidrule(lr){3-4}
\cmidrule(lr){5-5}
& & \textbf{TP} & \textbf{TR} & \textbf{Others} \\
\midrule

\multirow{8}{*}{LongVA-7B}
& \textsc{Base}
& 60.0 & 40.1 & 55.0 \\

& \textsc{LoRA}$^{\ddagger}$
& \traincell{72.7}
& \traincell{53.1}
& 54.7 \\

& \textsc{FFT}$^{\ddagger}$
& \traincell{81.8}
& \traincell{64.4}
& 53.0 \\

& \textsc{Full-backbone}
& 60.0 & 40.1 & 55.0 \\

& \textsc{All-attention}
& \textbf{61.8} & 39.5 & 55.2 \\

& \textsc{Random-k}
& \multicolumn{1}{c}{58.9$\pm$1.6}
& \multicolumn{1}{c}{38.4$\pm$1.7}
& \multicolumn{1}{c}{54.7$\pm$0.5} \\

& \textsc{Random-k}$^{\dagger}$
& \multicolumn{1}{c}{59.6$\pm$0.8}
& \multicolumn{1}{c}{40.8$\pm$0.9}
& \multicolumn{1}{c}{54.8$\pm$0.5} \\
\cmidrule(lr){2-5}

& \multicolumn{1}{l}{\ourcell\textbf{\ours{}}}
& {\ourcell \textbf{61.8} \deltaup{(+3.0\%)}}
& {\ourcell \textbf{49.7} \deltaup{(+23.9\%)}}
& {\ourcell \textbf{55.3} \deltaup{(+0.5\%)}} \\

\midrule

\multirow{8}{*}{InternVL3-8B}
& \textsc{Base}
& 74.5 & 52.0 & 67.4 \\

& \textsc{LoRA}$^{\ddagger}$
& \traincell{83.6}
& \traincell{71.2}
& 67.4 \\

& \textsc{FFT}$^{\ddagger}$
& \traincell{85.5}
& \traincell{71.2}
& 66.3 \\

& \textsc{Full-backbone}
& 76.4 & 53.1 & 67.3 \\

& \textsc{All-attention}
& 76.4 & 54.2 & 67.3 \\

& \textsc{Random-k}
& \multicolumn{1}{c}{78.2$\pm$2.9}
& \multicolumn{1}{c}{51.6$\pm$0.9}
& \multicolumn{1}{c}{67.5$\pm$0.3} \\

& \textsc{Random-k}$^{\dagger}$
& \multicolumn{1}{c}{78.6$\pm$2.3}
& \multicolumn{1}{c}{53.1$\pm$0.9}
& \multicolumn{1}{c}{67.3$\pm$0.3} \\
\cmidrule(lr){2-5}

& \multicolumn{1}{l}{\ourcell\textbf{\ours{}}}
& {\ourcell \textbf{81.8} \deltaup{(+9.8\%)}}
& {\ourcell \textbf{57.6} \deltaup{(+10.8\%)}}
& {\ourcell \textbf{67.9} \deltaup{(+0.7\%)}} \\

\midrule

\multirow{8}{*}{Qwen3-VL-4B}
& \textsc{Base}
& 69.1 & 45.2 & 65.3 \\

& \textsc{LoRA}$^{\ddagger}$
& \traincell{76.4}
& \traincell{57.1}
& 65.0 \\

& \textsc{FFT}$^{\ddagger}$
& \traincell{85.5}
& \traincell{70.6}
& 64.1 \\

& \textsc{Full-backbone}
& 67.3 & 45.8 & 64.7 \\

& \textsc{All-attention}
& 67.3 & 44.1 & \textbf{65.5} \\

& \textsc{Random-k}
& \multicolumn{1}{c}{63.3$\pm$3.0}
& \multicolumn{1}{c}{41.5$\pm$2.7}
& \multicolumn{1}{c}{62.7$\pm$1.2} \\

& \textsc{Random-k}$^{\dagger}$
& \multicolumn{1}{c}{69.5$\pm$1.5}
& \multicolumn{1}{c}{44.4$\pm$1.1}
& \multicolumn{1}{c}{62.4$\pm$2.6} \\
\cmidrule(lr){2-5}

& \multicolumn{1}{l}{\ourcell\textbf{\ours{}}}
& {\ourcell \textbf{70.9} \deltaup{(+2.6\%)}}
& {\ourcell \textbf{46.9} \deltaup{(+3.8\%)}}
& {\ourcell 64.6 \deltadown{(-1.1\%)}} \\

\bottomrule
\end{tabular}
\end{adjustbox}
\label{tab:video_mme_tp}
\vspace{-1em}
\end{wraptable}

\vspace{-0.3em}
\subsection{TR Recovery Generalizes Beyond the Diagnostic Set}
\label{sec:main_results}
\vspace{-0.3em}

Tables~\ref{tab:video_mme_tp} and~\ref{tab:other_benchmarks} test whether adaptation-induced TR degradation can be recovered without sacrificing the perceptual capability acquired during multimodal adaptation.

\textbf{Recovery transfers without undoing multimodal adaptation.}
On Video-MME, \ours{} improves TR by 23.9\%, 10.8\%, and 3.8\% for LongVA-7B, InternVL3-8B, and Qwen3-VL-4B, respectively, while preserving or improving TP in all three models. The held-out \textsc{Others} subset changes by only $+0.5\%$, $+0.7\%$, and $-1.1\%$, indicating limited impact on non-target capabilities. More importantly, after selection on only 232 Video-MME examples, each recipe remains fixed and improves reasoning-aligned performance over \textsc{Base} in all 12 comparisons across four unseen benchmarks. Including Video-MME TR, \ours{} improves all 15 reasoning-aligned evaluations, with relative gains reaching 27.8\% on MMBench-Video TR and 24.3\% on Video-Holmes. Thus, the selected interventions recover reasoning behavior that transfers across datasets without broadly erasing the capabilities acquired during adaptation.

\textbf{The same recovery appears at the language-backbone level.}
On TempReason, multimodal adaptation reduces text-only TR across all three model families, and \ours{} consistently restores part of this loss (\S~\ref{app:tempreason}). This closes the loop between diagnosis and recovery: the same intervention that improves video reasoning also repairs the text-only temporal competence weakened during adaptation. An additional evaluation on InternVL3-14B shows the same recovery pattern beyond the 4B--8B setting (\S~\ref{app:scalability}). Selected recipes are reported in \S~\ref{app:recipes-stability}.

\vspace{-0.3em}
\begin{table*}[t]
\vspace{-0.5em}
\centering
\caption{\textbf{\ours{} transfers temporal reasoning gains beyond the diagnostic set.}
Fixed recipes selected on Video-MME are evaluated on four unseen video benchmarks. Parentheses show relative change over \textsc{Base}; $\dagger$ denotes $\alpha$-tuned \textsc{Random-k}. MMBench-Video uses an LLM-as-a-judge score; all other results are accuracy (\%). Best performance is marked in bold.}
\vspace{0.3em}
\small
\renewcommand{\arraystretch}{1.08}
\setlength{\tabcolsep}{4pt}

\begin{adjustbox}{max width=\textwidth}
\begin{tabular}{l l
                S[table-format=2.1]
                S[table-format=2.1]
                S[table-format=2.1]
                S[table-format=2.1]
                S[table-format=1.2]
                S[table-format=1.2]
                S[table-format=2.1]}
\toprule
\textbf{Model} & \textbf{Variant}
& \multicolumn{2}{c}{\textbf{LongVideoBench}}
& \multicolumn{2}{c}{\textbf{LVBench}}
& \multicolumn{2}{c}{\textbf{MMBench-Video}}
& \multicolumn{1}{c}{\textbf{Video-Holmes}} \\
\cmidrule(lr){3-4}
\cmidrule(lr){5-6}
\cmidrule(lr){7-8}
\cmidrule(lr){9-9}
& & {\textbf{Relation}} & {\textbf{Overall}}
  & {\textbf{TR}} & \textbf{{Overall}}
  & {\textbf{TR}} & {\textbf{Overall}}
  & {\textbf{Overall}} \\
  
\midrule

\multirow[c]{8}{*}[-0.7ex]{LongVA-7B}
& \textsc{Base}
& 48.0 & 52.5
& 45.3 & 37.6
& 1.05 & 1.27
& 15.2 \\

& \textsc{LoRA}
& 47.2 & 52.2
& 45.3 & 37.7
& 1.04 & 1.24
& 14.5 \\

& \textsc{FFT}
& 35.5 & 40.8
& 41.8 & 35.4
& 0.98 & 1.15
& 4.7 \\

& \textsc{Full-backbone}
& 47.9 & 52.9
& 44.8 & 38.1
& 1.08 & 1.25
& 14.6 \\

& \textsc{All-attention}
& 48.0 & 52.6
& 46.3 & 38.0
& 1.08 & 1.28
& 15.4 \\

& \textsc{Random-k}
& \multicolumn{1}{c}{48.1$\pm$0.7}
& \multicolumn{1}{c}{52.7$\pm$0.2}
& \multicolumn{1}{c}{44.4$\pm$1.6}
& \multicolumn{1}{c}{37.8$\pm$0.6}
& \multicolumn{1}{c}{1.07$\pm$0.02}
& \multicolumn{1}{c}{1.25$\pm$0.02}
& \multicolumn{1}{c}{14.5$\pm$3.8} \\

& \textsc{Random-k}$^{\dagger}$
& \multicolumn{1}{c}{48.2$\pm$0.7}
& \multicolumn{1}{c}{52.9$\pm$0.3}
& \multicolumn{1}{c}{45.3$\pm$1.8}
& \multicolumn{1}{c}{37.7$\pm$0.8}
& \multicolumn{1}{c}{1.07$\pm$0.01}
& \multicolumn{1}{c}{1.25$\pm$0.02}
& \multicolumn{1}{c}{15.8$\pm$2.5} \\
\cmidrule(lr){2-9}

& \multicolumn{1}{l}{\ourcell\textbf{\ours{}}}
& {\ourcell \textbf{48.6} \deltaup{(+1.3\%)}}
& {\ourcell \textbf{53.6} \deltaup{(+2.1\%)}}
& {\ourcell \textbf{46.8} \deltaup{(+3.3\%)}}
& {\ourcell \textbf{39.7} \deltaup{(+5.6\%)}}
& {\ourcell \textbf{1.09} \deltaup{(+3.8\%)}}
& {\ourcell \textbf{1.29} \deltaup{(+1.6\%)}}
& {\ourcell \textbf{18.9} \deltaup{(+24.3\%)}} \\

\midrule

\multirow[c]{8}{*}[-0.7ex]{InternVL3-8B}
& \textsc{Base}
& 54.8 & 59.8
& 51.2 & 44.8
& 1.44 & 1.68
& 36.5 \\

& \textsc{LoRA}
& 54.4 & 58.9
& 52.2 & 44.7
& 1.42 & 1.71
& 36.2 \\

& \textsc{FFT}
& 52.2 & 55.7
& 50.2 & 44.4
& 1.41 & 1.69
& 35.7 \\

& \textsc{Full-backbone}
& 55.1 & 59.9
& 50.7 & 44.7
& 1.44 & 1.71
& 36.3 \\

& \textsc{All-attention}
& 54.8 & 59.5
& 50.2 & 44.7
& 1.44 & 1.71
& 36.2 \\

& \textsc{Random-k}
& \multicolumn{1}{c}{55.1$\pm$0.8}
& \multicolumn{1}{c}{59.5$\pm$0.5}
& \multicolumn{1}{c}{49.8$\pm$1.3}
& \multicolumn{1}{c}{44.2$\pm$0.7}
& \multicolumn{1}{c}{1.43$\pm$0.02}
& \multicolumn{1}{c}{1.70$\pm$0.01}
& \multicolumn{1}{c}{36.7$\pm$1.0} \\

& \textsc{Random-k}$^{\dagger}$
& \multicolumn{1}{c}{\textbf{55.2$\pm$0.5}}
& \multicolumn{1}{c}{60.0$\pm$0.2}
& \multicolumn{1}{c}{48.1$\pm$0.8}
& \multicolumn{1}{c}{44.3$\pm$0.3}
& \multicolumn{1}{c}{1.44$\pm$0.02}
& \multicolumn{1}{c}{1.70$\pm$0.01}
& \multicolumn{1}{c}{36.5$\pm$0.5} \\
\cmidrule(lr){2-9}

& \multicolumn{1}{l}{\ourcell\textbf{\ours{}}}
& {\ourcell \textbf{55.2} \deltaup{(+0.7\%)}}
& {\ourcell \textbf{60.3} \deltaup{(+0.8\%)}}
& {\ourcell \textbf{53.2} \deltaup{(+3.9\%)}}
& {\ourcell \textbf{45.0} \deltaup{(+0.4\%)}}
& {\ourcell \textbf{1.46} \deltaup{(+1.4\%)}}
& {\ourcell \textbf{1.72} \deltaup{(+2.4\%)}}
& {\ourcell \textbf{37.2} \deltaup{(+1.9\%)}} \\

\midrule

\multirow[c]{8}{*}[-0.7ex]{Qwen3-VL-4B}
& \textsc{Base}
& 56.0 & 61.9
& 39.3 & 39.4
& 1.08 & 1.50
& 31.5 \\

& \textsc{LoRA}
& 55.5 & 61.1
& 42.3 & 40.4
& 1.24 & 1.56
& 32.0 \\

& \textsc{FFT}
& 51.5 & 55.3
& 43.3 & 40.3
& 1.11 & 1.53
& 30.7 \\

& \textsc{Full-backbone}
& 55.9 & 60.4
& 33.3 & 34.9
& 1.26 & 1.56
& 31.2 \\

& \textsc{All-attention}
& 56.3 & \textbf{62.0}
& 43.3 & \textbf{43.7}
& 1.27 & 1.59
& 31.9 \\

& \textsc{Random-k}
& \multicolumn{1}{c}{55.2$\pm$0.8}
& \multicolumn{1}{c}{59.8$\pm$0.4}
& \multicolumn{1}{c}{40.5$\pm$2.9}
& \multicolumn{1}{c}{39.0$\pm$1.5}
& \multicolumn{1}{c}{1.27$\pm$0.11}
& \multicolumn{1}{c}{1.55$\pm$0.08}
& \multicolumn{1}{c}{31.0$\pm$0.8} \\

& \textsc{Random-k}$^{\dagger}$
& \multicolumn{1}{c}{56.2$\pm$0.3}
& \multicolumn{1}{c}{60.6$\pm$0.4}
& \multicolumn{1}{c}{41.5$\pm$1.4}
& \multicolumn{1}{c}{40.0$\pm$1.0}
& \multicolumn{1}{c}{1.26$\pm$0.08}
& \multicolumn{1}{c}{1.55$\pm$0.11}
& \multicolumn{1}{c}{31.7$\pm$0.7} \\
\cmidrule(lr){2-9}

& \multicolumn{1}{l}{\ourcell\textbf{\ours{}}}
& {\ourcell \textbf{56.9} \deltaup{(+1.6\%)}}
& {\ourcell 60.6 \deltadown{(-2.1\%)}}
& {\ourcell \textbf{46.3} \deltaup{(+17.8\%)}}
& {\ourcell 40.6 \deltaup{(+3.0\%)}}
& {\ourcell \textbf{1.38} \deltaup{(+27.8\%)}}
& {\ourcell \textbf{1.66} \deltaup{(+10.7\%)}}
& {\ourcell \textbf{32.9} \deltaup{(+4.4\%)}} \\

\bottomrule
\end{tabular}
\end{adjustbox}
\label{tab:other_benchmarks}
\vspace{-10pt}
\end{table*}

\subsection{Successful Recovery Depends on Where and How Parameters Are Restored}
\label{sec:design_validation}
\vspace{-0.3em}

We next ask what makes recovery transferable, isolating the roles of intervention structure, search strategy, and supervised fitting.

\textbf{Recovery depends on both where and what is restored.}
As shown in Tables~\ref{tab:video_mme_tp} and~\ref{tab:other_benchmarks}, we tune $\alpha$ independently for \textsc{Full-Backbone}, \textsc{All-Attention}, and \textsc{Random-k} on the same TP/TR diagnostic set used by \ours{}. These baselines remain below \ours{} on reasoning-aligned evaluations even after controlling for interpolation strength, showing that the gains cannot be explained by a favorable mixing coefficient alone. We further fix the \ours{} layer pattern and compare attention-only merging with MLP-only, attention-plus-MLP, and full-block interventions. Table~\ref{tab:attention_mlp_ablation} shows that attention-only merging provides the strongest and most transferable recovery across Video-MME, LVBench, and Video-Holmes. Effective recovery therefore depends on both \emph{where} parameters are restored and \emph{which} parameters are modified within those layers.

\textbf{Structured search finds better recovery recipes.}
Figure~\ref{fig:merit_search_stability} compares CMA-ES with uniform random search under the same $\alpha$, recipe space, objective, and 1,600-candidate budget. CMA-ES reaches higher final objective values on all three VLMs, and its independent runs converge to the same final objective and recipe. On InternVL3-8B, Table~\ref{tab:search_strategy_comparison} further shows that the best random-search recipes improve Video-MME TR but remain below \ours{} on both the diagnostic task and unseen benchmarks. Structured exploration of dependencies among layer assignments therefore matters even under a matched search space and evaluation budget.

\textbf{Better fitting does not imply better recovery.}
Tables~\ref{tab:video_mme_tp} and~\ref{tab:other_benchmarks} show that \textsc{LoRA} and \textsc{FFT} achieve substantially higher TP/TR accuracy on the 232 optimization examples, yet their gains transfer less reliably to held-out categories and unseen benchmarks. \ours{} can only recombine parameters already present in the paired checkpoints. This gap between in-sample fit and out-of-sample transfer is consistent with \ours{} selecting reusable competence already encoded in the model, while supervised tuning can specialize more strongly to the diagnostic set.

A fixed-$\alpha$ search on InternVL3-8B evaluates at most 1,600 candidates and takes approximately 1.7 hours using 4 GPUs and 16 parallel workers (Table~\ref{tab:cost_efficiency}). As described in \S~\ref{subsec:search}, the complete grid contains at most 8,001 candidate evaluations ($5\times1{,}600$ search evaluations plus one evaluation at $\alpha=0.5$), and the five searches can be parallelized across $\alpha$. Once selected, the recipe is compiled into the checkpoint and adds no trainable parameters or inference overhead. Full search comparisons and cost details are provided in \S~\ref{app:recipes-stability} and \S~\ref{app:efficiency}.

\begin{figure}[t]
  \centering
  \includegraphics[width=0.85\columnwidth]{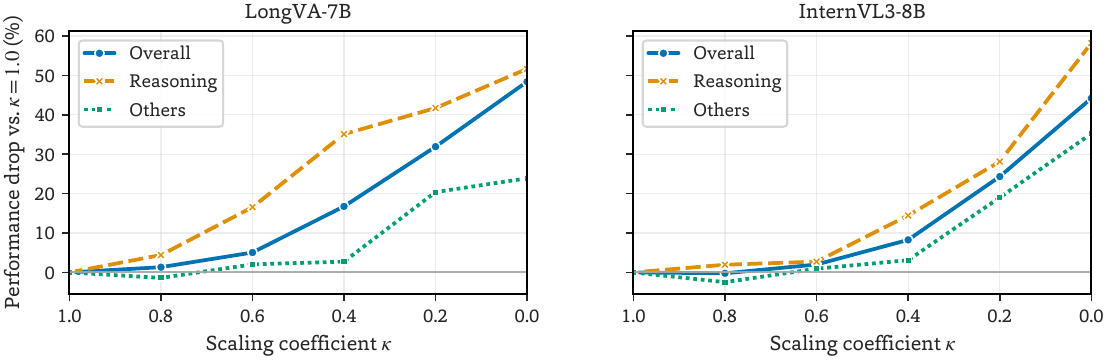}
\caption{\textbf{\ours{} selects reasoning-sensitive intervention points.}
Suppressing the self-attention outputs of the selected layers in the unmodified VLM causes \textit{Reasoning} to degrade markedly faster than \textit{Overall} and \textit{Others}.}
\vspace{-0.5em}
\label{fig:masking}
\end{figure}

\subsection{Selected Layers Are Reasoning-Sensitive Intervention Points}
\label{sec:masking}

The previous results show that layer identity matters for recovery. We next ask whether the selected locations also have a distinct functional relationship with temporal reasoning. To separate this question from merging itself, we intervene directly on the unmodified VLM by suppressing the self-attention outputs of the selected layers.

For each selected layer, we scale its self-attention output by $\kappa\in[0,1]$, where $\kappa=1$ leaves the output unchanged and $\kappa=0$ fully suppresses it while preserving the residual connection. We evaluate LVBench at $\kappa\in\{1.0,0.8,0.6,0.4,0.2,0.0\}$ and define
\begin{equation}
\Delta(\kappa)
=
\frac{\mathrm{Score}(1.0)-\mathrm{Score}(\kappa)}
{\mathrm{Score}(1.0)}.
\end{equation}
We compare \textbf{Reasoning}, \textbf{Overall}, and \textbf{Others}, where \textbf{Others} is the mean of all non-reasoning sub-metrics.

Figure~\ref{fig:masking} shows a clear separation. For LongVA-7B at $\kappa=0.4$, \textbf{Reasoning} drops by 35.1\%, compared with 16.8\% for \textbf{Overall} and 2.7\% for \textbf{Others}; complete suppression raises the reasoning drop to 51.7\%. InternVL3-8B exhibits the same pattern: at $\kappa=0$, the corresponding drops are 58.2\%, 44.2\%, and 35.2\%. The layers discovered through recovery search therefore coincide with intervention points to which temporal reasoning is especially sensitive.

\begin{figure*}[t]
  \centering
  \includegraphics[width=0.95\textwidth]{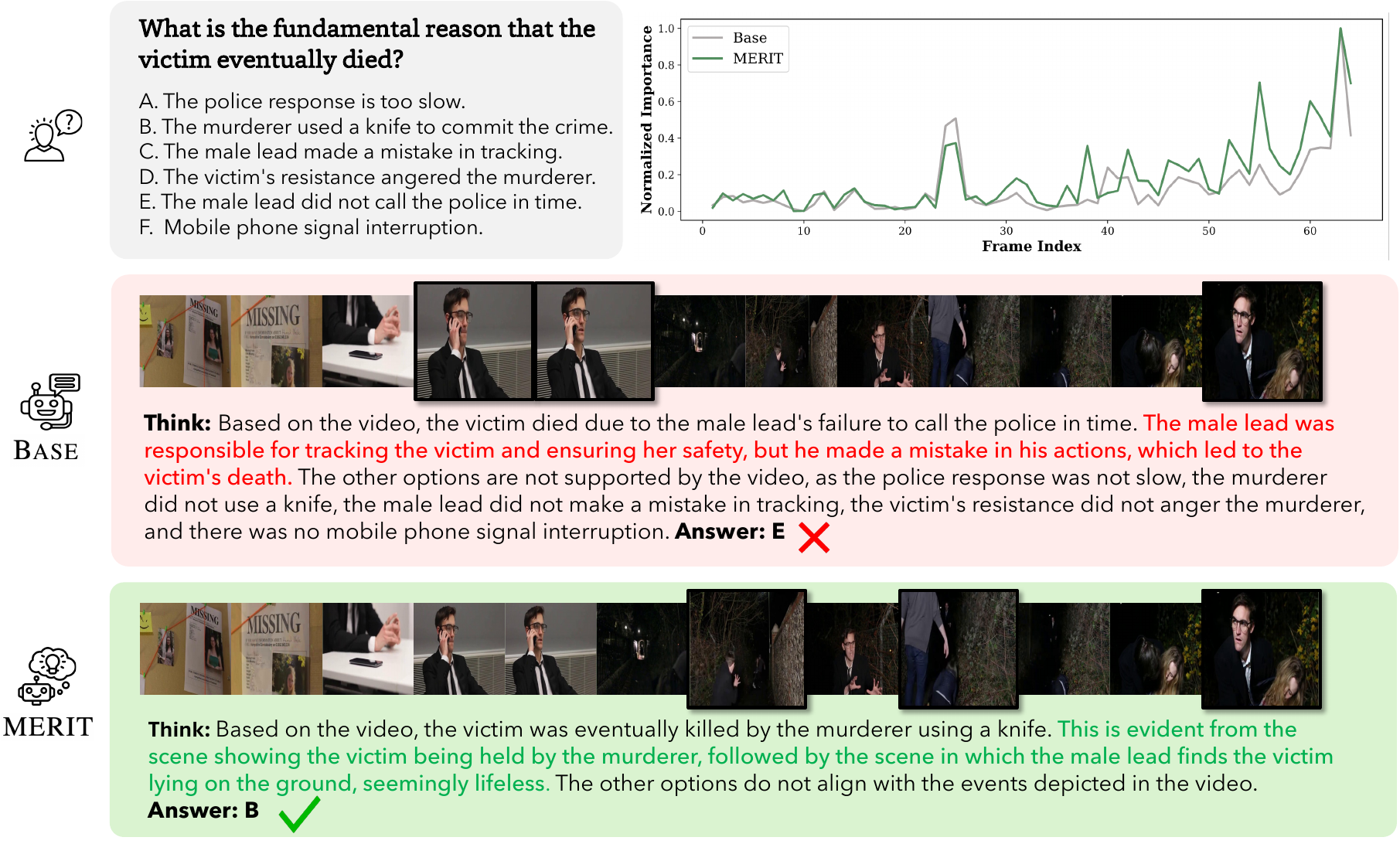}
\caption{\textbf{\ours{} shifts temporal grounding from a salient local cue to distributed causal evidence.}
In this Video-Holmes example (Question ID: 463), \textsc{Base} anchors on the phone-call scene and incorrectly attributes the victim's death to delayed help. \ours{} instead assigns greater support to two temporally separated events, the victim being restrained and later found motionless, and recovers the correct causal explanation.}
\vspace{-0.5em}
\label{fig:example1}
\end{figure*}

\subsection{Qualitative Analysis: Recovery Changes Temporal Grounding}
\label{sec:attribution}

Accuracy alone does not reveal whether recovery changes the evidence underlying a model's decision. We therefore examine how \ours{} redistributes visual support across time. Using temporal-reasoning questions from Video-Holmes and InternVL3-8B as a representative model, we combine frame-level attribution at the answer step with generated reasoning traces.

\textbf{Attributing the final decision.}
For an input $x$ and generated output $y$ formatted as \texttt{<think>...</think>\textbackslash n<answer>...</answer>}, we compute attribution at the prediction step for the answer-option token. Let $t$ denote this token's output position, $y_{<t}$ the generated prefix ending immediately before it, and $z_t$ the next-token logits obtained under teacher forcing on $(x,y_{<t})$. Let $\mathcal{C}$ denote the set of candidate answer-option tokens and $c\in\mathcal{C}$ the model-selected option. Restricting $z_t$ to $\mathcal{C}$, we define the attribution target as the logit margin between $c$ and its strongest alternative:
\begin{equation}
s =
z_t^{\mathcal{C}}(c)
-
\max_{c'\in\mathcal{C}\setminus\{c\}}
z_t^{\mathcal{C}}(c').
\end{equation}

\textbf{Frame-level evidence.}
Suppose the input contains $T$ sampled frames. Let $V_{\tau}\in\mathbb{R}^{K\times d}$ denote the $K$ projected visual-token embeddings of frame $\tau$. We aggregate absolute gradient--activation products to obtain a non-negative attribution score for each frame:
\begin{equation}
I_{\tau} =
\sum_{k=1}^{K}
\sum_{j=1}^{d}
\left|
\frac{\partial s}{\partial V_{\tau}[k,j]}
\cdot
V_{\tau}[k,j]
\right|.
\end{equation}
For each model and input, we min--max normalize $\{I_{\tau}\}_{\tau=1}^{T}$ across frames to obtain a relative temporal attribution profile.

\textbf{Recovery can shift support beyond salient local cues.}
Figure~\ref{fig:example1} shows a representative Video-Holmes example. The base model concentrates on a salient phone-call scene and incorrectly attributes the victim's death to delayed help. \ours{} instead assigns greater importance to two temporally separated events: the victim being restrained and later found motionless. Linking these observations supports the correct causal explanation even though the killing itself is not explicitly shown.

This case illustrates a qualitative change beyond answer accuracy: the repaired model shifts support away from a misleading local cue and toward temporally separated events that jointly determine the causal explanation. Additional examples in \S~\ref{sec:appendix_examples} show related changes, including cases where both models reach the same answer but \ours{} grounds it in a longer temporal event structure.

\vspace{-0.3em}
\section{Conclusion}
\vspace{-0.3em}

We study temporal reasoning degradation in VLMs as a capability recovery problem. \ours{} selectively reuses the paired pre-adaptation language checkpoint to restore reasoning while preserving multimodal perception. Across three VLM families, compactly selected recipes transfer to four unseen benchmarks, and controlled analyses show that recovery depends on both where and what parameters are restored. Interventional analysis further links the selected layers to temporal reasoning sensitivity, and qualitative attribution reveals corresponding changes in temporal evidence use. More broadly, multimodal adaptation need not permanently erase prior competence: when the original capability remains encoded in a paired checkpoint, targeted parameter recovery can restore it while retaining what adaptation added.

\clearpage


\subsection*{AI Use Statement}

Generative AI tools were used for language polishing and limited feedback on the presentation and organization of the manuscript. All research conceptualization, methodological decisions, experimental execution, validation, analysis, and final scientific judgments were made by the authors. Separately, pretrained language and multimodal models were used as components of the experimental methodology, including LLM-based evaluation following official benchmark protocols, as described in the paper. The authors reviewed all AI-assisted work and take full responsibility for the final content.

\subsection*{Ethics Statement}

This work uses existing models and benchmark datasets and does not involve human-subject studies or new data collection. We do not identify additional ethical concerns beyond those associated with the underlying models and datasets.

\subsection*{Reproducibility Statement}

We provide detailed experimental settings, model pairings, evaluation metrics, baseline configurations, implementation details, and search procedures in the appendix, together with complete results for all evaluated benchmarks.


\bibliography{iclr2027_conference}
\bibliographystyle{iclr2027_conference}


\clearpage
\appendix

\startcontents[appendix]

\section*{List of Appendices}
\printcontents[appendix]{}{1}{}
\clearpage

\section{Limitations and Future Work}

By framing adaptation-induced capability loss as a selective recovery problem, our study opens several directions beyond the current formulation.

First, \textbf{\ours{} currently relies on benchmark-defined signals for temporal perception and temporal reasoning.} This controlled separation makes recovery measurable, but assumes access to clean target and preservation signals. A natural next step is to replace them with weaker supervision, synthetic diagnostics, preference signals, or model-based evaluation. This would extend selective recovery to settings where degraded and protected capabilities are only partially observed.

Second, \textbf{the current parameterization deliberately favors a compact, tractable repair space.} \ours{} searches discrete layer-dominance patterns while selecting interpolation strength through an outer grid. Our finding that recovery depends on both where and what parameters are restored suggests a richer space that could jointly optimize layer selection, interpolation strength, intervention modules, or finer-grained structures. More sample-efficient black-box optimization could make such adaptive repair practical.

Finally, \textbf{model-specific recovery recipes offer a new lens on how capabilities are altered by adaptation.} Systematic comparisons across architectures and adaptation trajectories could reveal recurring intervention patterns, transferable search priors, or reusable rules for capability repair. Extending the same framework beyond temporal reasoning would further test which pretrained capabilities remain recoverable after multimodal adaptation. More broadly, this opens a path toward understanding when capability loss is reversible, where recoverable competence remains encoded, and how such recovery can be automated.

\section{Text-Only Recovery at the Language-Backbone Level}
\label{app:tempreason}

To test whether \ours{} repairs the language-side capability identified in Figure~\ref{fig:intro_pic}, we evaluate temporal reasoning without visual input. Comparing the original LLM, the adapted language backbone, and \ours{} isolates whether the recovery observed on video tasks extends to the temporal competence degraded within the language component itself.

\paragraph{Benchmark and Tasks.}
We evaluate on \textsc{TempReason} \citep{tan2023towards}, focusing on its two higher-level tasks: L2 time--event reasoning, which identifies the event associated with a queried time, and L3 event--event reasoning, which infers temporal relations between events. The evaluation contains 5,397 L2 and 4,426 L3 examples.

\paragraph{Evaluation Protocol.}
Within each model family, the original LLM, adapted language backbone (\textsc{VLM Text-Only}), and \ours{} use the same prompt, with each model formatted using its native tokenizer and chat template:
\begin{quote}
\small
\texttt{Answer the question using only the temporal facts in the context.}\\
\texttt{Return only a short answer, without explanation.}\\[0.3em]
\texttt{Context: \{context\}}\\
\texttt{Question: \{question\}}\\
\texttt{Answer:}
\end{quote}
Decoding is deterministic with temperature $0$ and stops at the first newline. We report lenient Exact Match (EM, \%): after normalization, a prediction is correct if it exactly matches or contains a reference answer, accommodating minor wrappers produced by chat models.

\begin{figure}[t]
\centering
\includegraphics[width=\columnwidth]{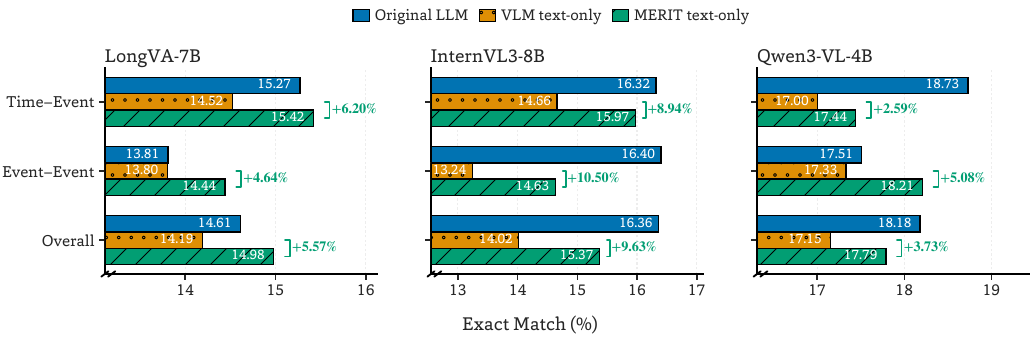}
\caption{\textbf{\ours{} recovers temporal reasoning weakened within the adapted language backbone.}
Across all three model families, multimodal adaptation reduces text-only \textsc{TempReason} performance, while \ours{} restores part or all of the lost capability. Percentages beside the brackets show relative gains over the adapted backbone.}
\label{fig:tempreason_merit}
\end{figure}

\paragraph{Backbone-Level Recovery.}
Figure~\ref{fig:tempreason_merit} shows a consistent pattern across all three model families: multimodal adaptation weakens text-only TR, and \ours{} recovers part or all of the lost performance. For LongVA-7B, \ours{} improves the adapted backbone by 6.20\%, 4.64\%, and 5.57\% on time--event, event--event, and overall EM, respectively, surpassing the original LLM on all three metrics. InternVL3-8B exhibits the largest degradation, especially on event--event reasoning, where EM falls from 16.40 to 13.24; \ours{} raises it to 14.63 and improves overall EM by 9.63\%. Qwen3-VL-4B follows the same pattern, with overall EM decreasing from 18.18 to 17.15 after adaptation and recovering to 17.79 with \ours{}. These results close the loop between diagnosis and recovery: the intervention that improves video reasoning also restores temporal competence within the adapted language backbone.

\section{Experimental Details}
\label{app:exp-details}

\subsection{Model Pairings}
\label{sec:model_details}

We pair each VLM with the pre-adaptation language backbone specified by its original release: LongVA-7B with Qwen2-7B-Instruct \citep{yang2024qwen2technicalreport}, InternVL3-8B with Qwen2.5-7B-Instruct \citep{qwen2025qwen25technicalreport}, and Qwen3-VL-4B-Instruct with Qwen3-4B-Instruct-2507 \citep{yang2025qwen3}. LongVA-7B and InternVL3-8B contain 28 self-attention layers, while Qwen3-VL-4B contains 36.

\subsection{Evaluation Tasks and Metrics}
\label{sec:metric_details}

We report each benchmark's metric most closely aligned with temporal reasoning: Temporal Reasoning for Video-MME and MMBench-Video, Relation for LongVideoBench, and Reasoning for LVBench. For Video-Holmes, whose questions center on multi-step inference, we report overall accuracy. Dataset statistics and task formats are summarized in \S~\ref{sec:benchmark_details}.

\subsection{Baseline Construction}
\label{sec:baseline_details}

\textsc{Full-Backbone} interpolates the complete language backbone, while \textsc{All-Attention} interpolates every self-attention layer, following the parameter-level merging setting of \citet{chen2025bring}. \textsc{Random-k} uniformly samples a binary layer pattern with exactly $k$ self-attention layers assigned to the $\hat{u}_{\ell}=1$ branch of Eq.~\ref{eq:directional_merge}, where $k$ matches the number of layers assigned $\hat{u}_{\ell}^{\ast}=1$ in the corresponding \ours{} recipe; the remaining layers follow the $\hat{u}_{\ell}=0$ branch. We report mean and standard deviation over five runs. For the fixed-$\alpha$ \textsc{Random-k} baseline, we use the interpolation coefficient selected by \ours{} for the corresponding model. For $\alpha$-tuned baselines, the interpolation coefficient is selected on the same TP/TR diagnostic set used by \ours{}; for \textsc{Random-k}, $\alpha$ is tuned independently for each sampled layer pattern.

\textsc{LoRA} and full-parameter fine-tuning (\textsc{FFT}) use the same 232 diagnostic examples for training. \textsc{LoRA} applies rank-16 adapters to the language model with a learning rate of $4\times10^{-5}$, while \textsc{FFT} updates the full language backbone with a learning rate of $5\times10^{-6}$. Both use bfloat16, cosine scheduling, weight decay of $0.05$, and a warmup ratio of $0.03$. Training budgets and GPU usage are reported in Table~\ref{tab:cost_efficiency}.

\subsection{Implementation and Evaluation}
\label{sec:implementation_details}

Our implementation builds on \texttt{mergekit-evolve} \citep{goddard2024arcee}. For each fixed $\alpha$, CMA-ES uses a population size of
$\lambda_{\mathrm{pop}}=4+\lfloor 3\ln L\rfloor$
and evaluates at most 1,600 candidates under $\mathcal{F}$. We cache objective values for repeated discrete recipes and visual-processor outputs, yielding a 2--3$\times$ speedup. We set the perceptual-degradation weight to $\lambda=55/177$, matching the sizes of the TP and TR diagnostic subsets so that a one-example decrease in TP accuracy carries the same weight as a one-example increase in TR accuracy.

We evaluate Video-MME, LongVideoBench, and LVBench with \texttt{lmms-eval} \citep{zhang2025lmms}, and MMBench-Video and Video-Holmes with VLMEvalKit \citep{duan2024vlmevalkit}. Following the official benchmark implementations \citep{fang2024mmbench,cheng2025video}, MMBench-Video and Video-Holmes use GPT-4-Turbo and GPT-4-0125 \citep{achiam2023gpt} as judges. Decoding is deterministic with temperature $0$. We use at most 128 frames for LongVA and 64 frames for InternVL3 and Qwen3-VL; for Qwen3-VL, we set \texttt{min\_pixels}=8,192 and \texttt{max\_pixels}=786,432. MMBench-Video reports the mean judge score on the $\{0,1,2,3\}$ scale; all other benchmarks report accuracy.

\section{Additional Analyses of \ours{}}
\label{app:merit-additional}

This section provides complementary analyses of \ours{} beyond the main results. We examine the effectiveness and stability of evolutionary recipe search (\S~\ref{app:recipes-stability}), the choice of intervention module (\S~\ref{app:ablation}), search cost and deployment efficiency (\S~\ref{app:efficiency}), scaling to a larger VLM (\S~\ref{app:scalability}), and additional temporal-grounding case studies (\S~\ref{sec:appendix_examples}).

\subsection{Search Effectiveness and Recipe Stability}
\label{app:recipes-stability}

\paragraph{Recovered recipes are model-specific.}
Table~\ref{tab:appendix_merit_layers} reports the best recipe found for each VLM. The selected configurations differ in interpolation strength, number of LLM-dominant layers, and intervention locations. No single layer pattern therefore emerges across model families, motivating model-specific recipe search.

\begin{table*}[t]
\caption{\textbf{\ours{} discovers distinct recovery recipes across model families.}
We report the selected interpolation weight $\alpha$, number of LLM-dominant self-attention layers ($\hat{u}_{\ell}^{\ast}=1$), and their 0-based indices.}
\vspace{0.3em}
\centering
\footnotesize
\setlength{\tabcolsep}{5pt}
\renewcommand{\arraystretch}{1.05}

\begin{tabularx}{0.90\textwidth}{
    @{}
    l
    c
    c
    >{\raggedright\arraybackslash}X
    @{}
}
\toprule
\textbf{VLM}
& $\boldsymbol{\alpha}$
& \textbf{\#Layers}
& $\boldsymbol{\hat{u}_{\ell}^{\ast}=1}$ \textbf{ self-attention layer indices} \\
\midrule
LongVA-7B
& 1.0
& 11
& 1, 5, 7--8, 11, 13--14, 16, 19--20, 23 \\

InternVL3-8B
& 0.9
& 9
& 4, 7, 12, 14, 16--18, 24, 26 \\

Qwen3-VL-4B
& 1.0
& 16
& 0, 4, 8, 10, 13, 17--18, 21, 23, 26, 28--32, 35 \\
\bottomrule
\end{tabularx}

\label{tab:appendix_merit_layers}
\end{table*}

\paragraph{CMA-ES improves over matched-budget random search.}
We compare CMA-ES with uniform random search under the same $\alpha$, layer-recipe space, perception-constrained objective $\mathcal{F}(g)$, and 1,600-candidate budget. For each VLM, both methods use the $\alpha$ selected by \ours{} in Table~\ref{tab:appendix_merit_layers}; only the candidate-generation strategy differs. We run three independent trials for each method.

\begin{figure*}[t]
  \centering
  \includegraphics[width=\textwidth]{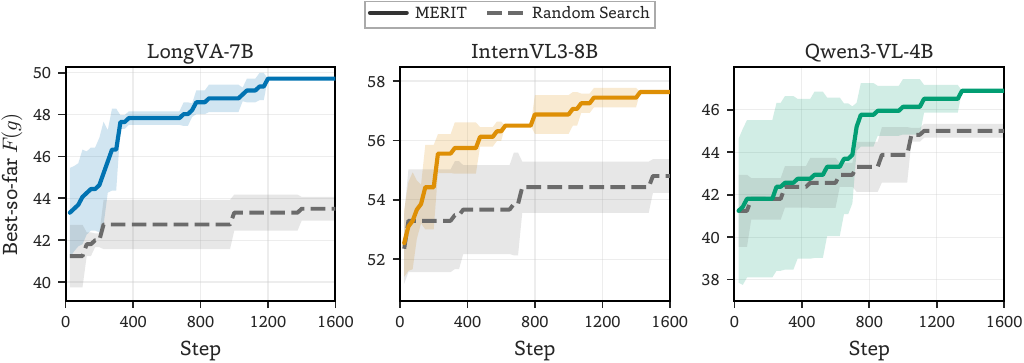}
  \caption{\textbf{CMA-ES outperforms matched-budget uniform random search.}
  Best-so-far perception-constrained objective $\mathcal{F}(g)$ over 1,600 candidate evaluations under the same $\alpha$, recipe space, objective, and budget. Curves show the mean over three runs; shaded regions indicate $\pm1$ standard deviation.}
  \label{fig:merit_search_stability}
\end{figure*}

Figure~\ref{fig:merit_search_stability} shows that CMA-ES reaches a higher mean final objective than random search on all three VLMs. Its three trials also recover the same final objective and recipe for each model despite different intermediate trajectories. This comparison complements the $\alpha$-tuned \textsc{Random-k} baseline, which separately controls for interpolation-strength selection across random layer sets.

\paragraph{The search advantage transfers downstream.}
For InternVL3-8B, we evaluate the best recipe from each random-search run on Video-MME, LVBench, and Video-Holmes using the same $\alpha=0.9$ as \ours{}. As shown in Table~\ref{tab:search_strategy_comparison}, random search improves Video-MME TR over \textsc{Base} but remains below \ours{} on both the diagnostic task and unseen benchmarks. The advantage of CMA-ES therefore extends beyond optimization of the diagnostic objective.

\begin{table*}[t]
\caption{\textbf{Downstream comparison of \ours{} and matched-budget
random search on InternVL3-8B.}
Both use $\alpha=0.9$. Random-search values report mean $\pm$ standard
deviation over the best recipes from three independent runs; the three
\ours{} runs recover the same final recipe and are therefore reported
once. All values are accuracy (\%).}
\vspace{0.3em}
\centering
\small
\renewcommand{\arraystretch}{1.08}
\setlength{\tabcolsep}{7pt}

\begin{adjustbox}{max width=\textwidth}
\begin{tabular}{l |
                c c |
                c c |
                c}
\toprule
Method
& \multicolumn{2}{c|}{Video-MME}
& \multicolumn{2}{c|}{LVBench}
& Video-Holmes \\
\cmidrule(lr){2-3}
\cmidrule(lr){4-5}
\cmidrule(lr){6-6}
& TR
& Overall
& TR
& Overall
& Overall \\
\midrule

\textsc{Base}
& 52.0
& 66.5
& 51.2
& 44.8
& 36.5 \\

\textsc{Random Search}
& $54.8{\pm}0.6$
& $66.4{\pm}0.4$
& $49.7{\pm}1.8$
& $43.8{\pm}0.6$
& $36.6{\pm}0.6$ \\

\ourcell\ours{}
& \ourcell{\bfseries 57.6}
& \ourcell{\bfseries 67.9}
& \ourcell{\bfseries 53.2}
& \ourcell{\bfseries 45.0}
& \ourcell{\bfseries 37.2} \\

\bottomrule
\end{tabular}
\end{adjustbox}
\label{tab:search_strategy_comparison}
\end{table*}

\subsection{Intervention-Module Ablations}
\label{app:ablation}

\paragraph{Interpolation weight under a fixed layer pattern.}
To test whether the recovery pattern found for self-attention transfers to other parameter groups, we fix the binary layer pattern $\hat{\mathbf{u}}^{\ast}$ discovered by \ours{} for InternVL3-8B and apply the directional interpolation in Eq.~\ref{eq:directional_merge} to MLP-only, attention-plus-MLP, and full-block parameters. For each variant, only $\alpha$ is tuned on the same 232-example TP/TR diagnostic set.

Here, the complement symmetry used by the original \ours{} search no longer removes coefficients below $0.5$, because the layer pattern is held fixed rather than complemented jointly with $\alpha$. We therefore sweep
$\alpha\in\{0.1,0.2,\ldots,1.0\}$.
Table~\ref{tab:attention_mlp_alpha_grid} reports the complete results.

\begin{table*}[t]
\caption{\textbf{Full interpolation-weight grid for the intervention-module ablation.} Each entry reports the TP/TR search objective $\mathcal{F}(g)$ on the 232-example diagnostic set. The binary layer pattern $\hat{\mathbf{u}}^{\ast}$ discovered by \ours{} is fixed across all variants, while the merged parameter group and interpolation coefficient $\alpha$ are varied. Bold indicates the selected $\alpha$ for each variant.}
\vspace{0.3em}
\centering
\small
\renewcommand{\arraystretch}{1.08}
\setlength{\tabcolsep}{4pt}
\sisetup{detect-weight=true, detect-inline-weight=math}

\begin{adjustbox}{max width=\textwidth}
\begin{tabular}{c | l |
                *{11}{S[table-format=2.1]}}
\toprule
Model & Variant
& {0.0} & {0.1} & {0.2} & {0.3} & {0.4} & {0.5}
& {0.6} & {0.7} & {0.8} & {0.9} & {1.0} \\
\midrule

\multirow[c]{3}{*}{InternVL3-8B}
& \textsc{MLP only}
& \multicolumn{1}{c}{--}
& {\bfseries 52.0}
& 51.4
& 51.4
& 51.4
& 50.3
& 50.8
& 49.7
& 46.3
& 46.9
& 45.8 \\

& \textsc{Attention + MLP}
& \multicolumn{1}{c}{--}
& 52.5
& 54.2
& {\bfseries 54.8}
& 54.2
& 53.7
& 52.5
& 52.0
& 50.3
& 49.7
& 46.9 \\

& \textsc{Full block}
& \multicolumn{1}{c}{--}
& 51.4
& 53.7
& {\bfseries 54.8}
& 54.2
& 54.2
& 53.1
& 52.0
& 50.8
& 49.7
& 47.5 \\

\bottomrule
\end{tabular}
\end{adjustbox}

\label{tab:attention_mlp_alpha_grid}
\end{table*}

The alternative modules favor substantially smaller interpolation weights: MLP-only selects $\alpha=0.1$, while attention-plus-MLP and full-block merging select $\alpha=0.3$. This shift indicates that the interpolation direction effective for self-attention does not transfer directly to MLP-related parameters.

\paragraph{Self-attention provides the strongest transferable recovery.}
Using the selected $\alpha$ for each variant, we evaluate the resulting models on Video-MME, LVBench, and Video-Holmes. Table~\ref{tab:attention_mlp_ablation} compares them with \textsc{Base} and the original attention-only \ours{} configuration.

\begin{table*}[t]
\caption{\textbf{Ablation over the intervention module on InternVL3-8B.} All non-base variants use the same fixed binary layer pattern $\hat{\mathbf{u}}^{\ast}$ discovered by \ours{}, with $\alpha$ tuned on the TP/TR diagnostic set. We report accuracy (\%) on the full Video-MME, LVBench, and Video-Holmes benchmarks. Best results are shown in bold.}
\vspace{0.3em}
\centering
\small
\renewcommand{\arraystretch}{1.08}
\setlength{\tabcolsep}{5pt}
\sisetup{detect-weight=true, detect-inline-weight=math}

\begin{adjustbox}{max width=\textwidth}
\begin{tabular}{l | c |
                S[table-format=2.1]
                S[table-format=2.1] |
                S[table-format=2.1]
                S[table-format=2.1] |
                S[table-format=2.1]}
\toprule
Variant
& {$\boldsymbol{\alpha}$}
& \multicolumn{2}{c|}{Video-MME}
& \multicolumn{2}{c|}{LVBench}
& {Video-Holmes} \\
\cmidrule(lr){3-4}
\cmidrule(lr){5-6}
\cmidrule(lr){7-7}
&
& {TR}
& {Overall}
& {TR}
& {Overall}
& {Overall} \\
\midrule

\textsc{Base}
& \multicolumn{1}{c|}{--}
& 52.0
& 66.5
& 51.2
& 44.8
& 36.5 \\

\ourcell\textsc{Attention only} (\ours{})
& \ourcell0.9
& \ourcell{\bfseries 57.6}
& \ourcell{\bfseries 67.5}
& \ourcell{\bfseries 53.2}
& \ourcell{\bfseries 45.0}
& \ourcell{\bfseries 37.2} \\

\textsc{MLP only}
& 0.1
& 52.0
& 66.5
& 49.8
& 44.6
& 36.5 \\

\textsc{Attention + MLP}
& 0.3
& 54.8
& 66.7
& 51.2
& 44.6
& 36.7 \\

\textsc{Full block}
& 0.3
& 54.8
& 66.7
& 50.7
& 44.7
& 36.7 \\

\bottomrule
\end{tabular}
\end{adjustbox}
\label{tab:attention_mlp_ablation}
\end{table*}

Attention-only merging provides the strongest and most consistent gains. It raises Video-MME TR from 52.0 to 57.6, LVBench Reasoning from 51.2 to 53.2, and Video-Holmes accuracy from 36.5 to 37.2. MLP-only merging provides no Video-MME TR gain and reduces LVBench Reasoning to 49.8, while attention-plus-MLP and full-block interventions remain below attention-only \ours{} on held-out evaluations. Thus, the selected layer pattern alone is insufficient: effective recovery also depends on which parameters are modified within those layers.

\subsection{Search Cost and Deployment Efficiency}
\label{app:efficiency}

\begin{table*}[t]
\caption{\textbf{Runtime and resource cost on InternVL3-8B.}
All measurements use NVIDIA A100-SXM4-80GB GPUs and the same 232-example TP/TR diagnostic set.
The \ours{} row reports one fixed-$\alpha$ search; full-grid cost is detailed in \S~\ref{app:efficiency}.}
\vspace{0.3em}
\centering
\small
\renewcommand{\arraystretch}{1.08}
\setlength{\tabcolsep}{4pt}

\begin{adjustbox}{max width=\textwidth}
\begin{tabular}{
    l
    l
    l
    l
    c
    l
    c
}
\toprule
\textbf{Method}
& \textbf{Data Usage}
& \textbf{Resources}
& \textbf{Optimization Budget}
& \textbf{Gradient Updates}
& \textbf{Trainable Parameters}
& \textbf{Wall-Clock Time} \\
\midrule

FFT
& 232 training examples
& 8 GPUs
& 1 epoch / 29 steps
& Yes
& 7.61B
& $\sim$0.5 h \\

LoRA
& 232 training examples
& 4 GPUs
& 3 epochs / 174 steps
& Yes
& 40.37M
& $\sim$1.6 h \\

One merging candidate
& 232 validation examples
& 1 GPU / 1 worker
& 1 candidate
& No
& 0
& $\sim$2 min \\

\ourcell \ours{} (per $\alpha$)
& \ourcell 232 search examples
& \ourcell 4 GPUs / 16 workers
& \ourcell 1,600 candidates
& \ourcell No
& \ourcell 0
& \ourcell $\sim$1.7 h \\

\bottomrule
\end{tabular}
\end{adjustbox}

\label{tab:cost_efficiency}
\end{table*}

Table~\ref{tab:cost_efficiency} reports computational cost on InternVL3-8B using NVIDIA A100-SXM4-80GB GPUs and the same 232-example TP/TR diagnostic set. \textsc{FFT} and \textsc{LoRA} use these examples for parameter optimization, whereas \ours{} uses them only as black-box signals for recipe selection.

Each fixed-$\alpha$ \ours{} search evaluates at most 1,600 candidates and takes approximately 1.7 hours using 4 GPUs and 16 parallel workers. As described in \S~\ref{subsec:search}, \ours{} searches five interpolation strengths independently, while $\alpha=0.5$ requires only one evaluation. The complete grid therefore contains at most 8,001 candidate evaluations ($5\times1{,}600$ search evaluations plus one evaluation at $\alpha=0.5$). Executed sequentially with the same 4-GPU allocation, the five searches require approximately 8.5 hours plus one single-candidate evaluation; they can instead be parallelized across $\alpha$.

All search cost is incurred offline. \ours{} performs no gradient updates, introduces no trainable parameters, and compiles the selected recipe directly into the checkpoint. The resulting model therefore retains the architecture and inference cost of the original VLM.

\subsection{Recovery Extends to a Larger VLM}
\label{app:scalability}

To test whether the recovery pattern extends beyond the 4B--8B models used in the main evaluation, we apply \ours{} to InternVL3-14B. As shown in Table~\ref{tab:internvl14b_scaling}, \ours{} improves Video-MME TR from 59.3 to 66.7, LVBench Reasoning from 48.8 to 50.7, and Video-Holmes accuracy from 37.3 to 38.0, while also improving overall Video-MME and LVBench performance. It outperforms full-backbone, all-attention, and $\alpha$-tuned \textsc{Random-k} merging on every reported metric. The same selective-recovery pattern therefore persists at a larger model scale.

\begin{table*}[t]
\caption{\textbf{Larger-scale validation on InternVL3-14B.}
\ours{} outperforms full-backbone, all-attention, and $\alpha$-tuned
\textsc{Random-k} interpolation across all reported reasoning-aligned
and overall metrics. Best performance in each evaluation column is
shown in bold. $\dagger$ denotes $\alpha$-tuned \textsc{Random-k}.}
\vspace{0.3em}
\centering
\small
\setlength{\tabcolsep}{4pt}
\renewcommand{\arraystretch}{1.1}

\begin{adjustbox}{max width=\textwidth}
\begin{tabular}{l l c c c c c c}
\toprule
\textbf{Model}
& \textbf{Variant}
& \textbf{Best $\boldsymbol{\alpha}$}
& \multicolumn{2}{c}{\textbf{Video-MME}}
& \multicolumn{2}{c}{\textbf{LVBench}}
& \multicolumn{1}{c}{\textbf{Video-Holmes}} \\
\cmidrule(lr){4-5}
\cmidrule(lr){6-7}
\cmidrule(lr){8-8}
& & 
& \textbf{TR}
& \textbf{Overall}
& \textbf{TR}
& \textbf{Overall}
& \textbf{Overall} \\
\midrule

\multirow[c]{5}{*}{InternVL3-14B}
& \textsc{Base}
& N/A
& 59.3
& 70.0
& 48.8
& 45.4
& 37.3 \\

& \textsc{Full-backbone}
& 0.1
& 60.5
& 69.9
& 49.3
& 45.9
& 37.4 \\

& \textsc{All-attention}
& 0.1
& 61.0
& 70.2
& 49.3
& 46.2
& 36.6 \\

& \textsc{Random-k}$^{\dagger}$
& 0.8, 0.6, 0.7, 0.0, 1.0
& $59.8{\pm}1.2$
& $70.0{\pm}0.2$
& $47.9{\pm}1.4$
& $45.9{\pm}0.4$
& $37.0{\pm}0.2$ \\

\cmidrule(lr){2-8}

& \ourcell \textbf{\ours{}}
& \ourcell 1.0
& \ourcell \textbf{66.7}
& \ourcell \textbf{70.6}
& \ourcell \textbf{50.7}
& \ourcell \textbf{46.9}
& \ourcell \textbf{38.0} \\

\bottomrule
\end{tabular}
\end{adjustbox}

\label{tab:internvl14b_scaling}
\end{table*}

\subsection{Additional Temporal-Grounding Case Studies}
\label{sec:appendix_examples}

We present three additional Video-Holmes cases using the same frame-level attribution and reasoning-trace analysis as in the main text. Together, they illustrate how recovery can alter temporal grounding both when it corrects an error and when the final answer is unchanged.

\begin{figure*}[t]
\centering
\includegraphics[width=\textwidth]{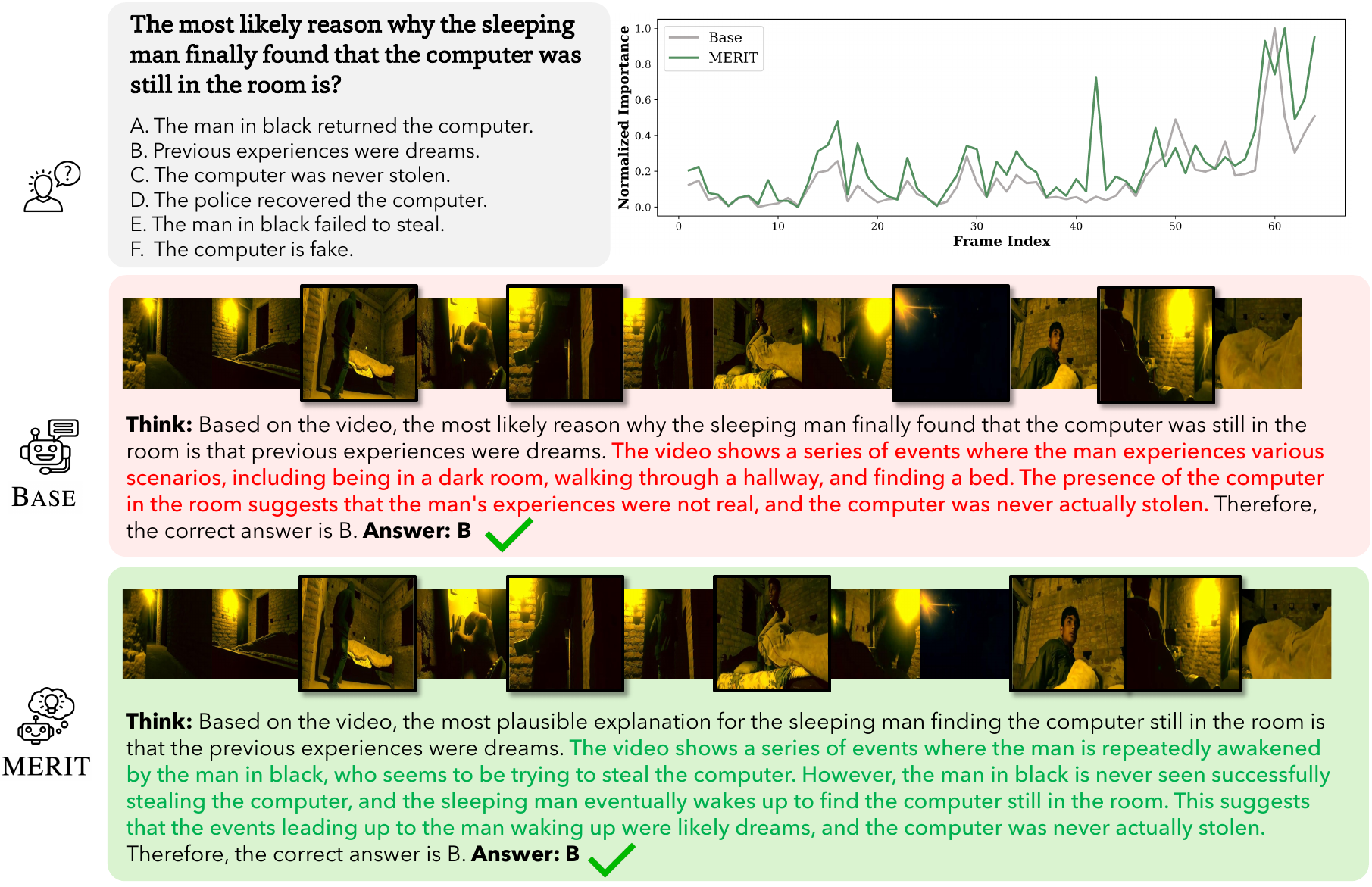}
\caption{\textbf{Case study from Video-Holmes (Question ID: 1040).} Both models answer correctly, but \ours{} grounds its prediction in a longer event chain linking the apparent theft, the pursuit, and the later awakening. This temporal structure supports the inference that the preceding events were a dream, while the base model remains more locally grounded.}
\label{fig:example2}
\end{figure*}

\begin{figure*}[!htbp]
\centering
\includegraphics[width=\textwidth]{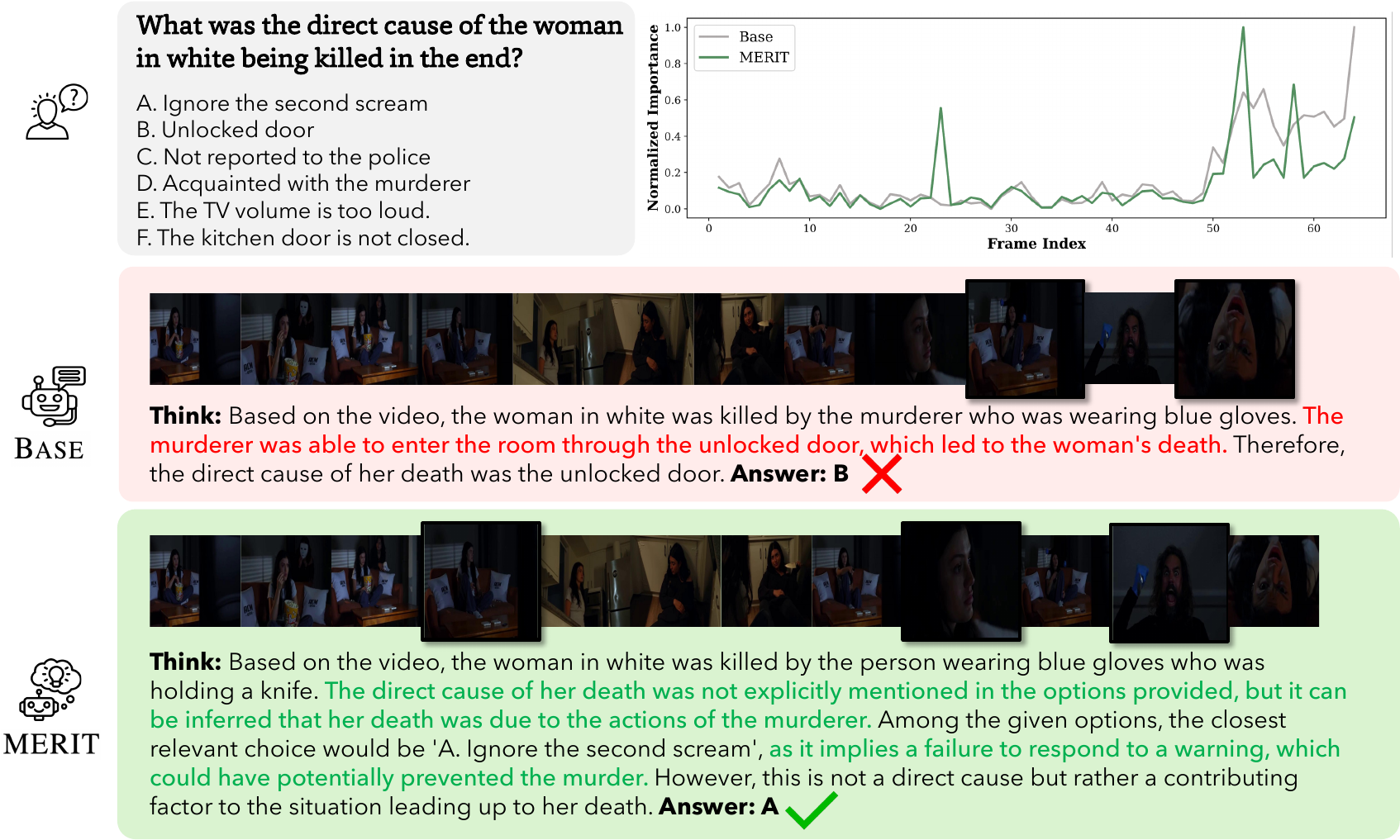}
\caption{\textbf{Case study from Video-Holmes (Question ID: 565).} The base model anchors on the murderer's entry through the unlocked door and incorrectly treats it as the direct cause of death. \ours{} instead links the staged scare, the first non-lethal encounter, and the ignored second scream, correctly identifying the missed warning as the decisive causal factor.}
\label{fig:example3}
\end{figure*}

\begin{figure*}[!htbp]
\centering
\includegraphics[width=\textwidth]{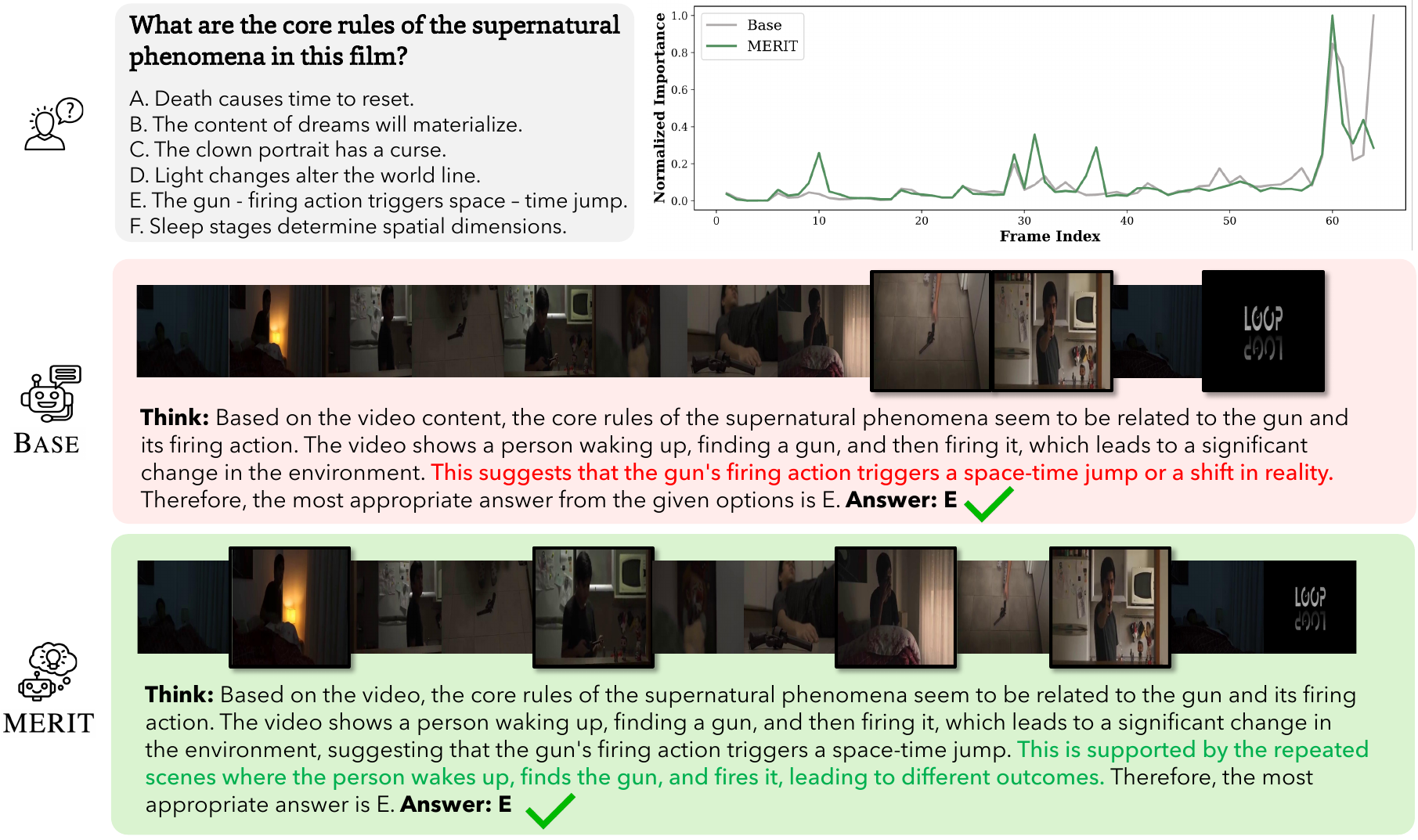}
\caption{\textbf{Case study from Video-Holmes (Question ID: 527).} Both models select the correct answer, but \ours{} links repeated wake-up and gun-firing sequences to infer a recurring pattern in which each gunshot triggers a shift in reality. The base model relies primarily on a single observed transition.}
\label{fig:example4}
\end{figure*}

In Figure~\ref{fig:example2}, both models answer correctly, but \ours{} links the apparent theft, subsequent pursuit, and later awakening into a longer event chain supporting the dream interpretation. In Figure~\ref{fig:example3}, \textsc{Base} anchors on the murderer's entry through an unlocked door, whereas \ours{} integrates the staged scare, repeated screams, and ignored warning to recover the correct causal explanation. In Figure~\ref{fig:example4}, both models again answer correctly, but \ours{} connects repeated wake-up and gun-firing sequences to infer a recurring temporal pattern. These examples show that recovery can change the temporal evidence structure underlying a prediction even when the final answer itself does not change.

\clearpage

\clearpage

\section{Interpolation-Controlled Merging Baselines}
\label{app:merging-baseline-details}

This section details how we construct the merging baselines used to isolate the effects of the interpolation coefficient and layer selection.

\paragraph{Controlling for interpolation strength.}
We independently tune $\alpha$ for \textsc{Full-Backbone}, \textsc{All-Attention}, and each \textsc{Random-k} baseline on the same 232-example TP/TR diagnostic set used by \ours{}. \textsc{Full-Backbone} and \textsc{All-Attention} use $\alpha\in\{0.1,0.2,\ldots,1.0\}$, with $\alpha=0$ omitted because it reduces to the unmodified VLM. For \textsc{Random-k}, each run fixes a randomly sampled binary layer pattern with exactly $k$ layers assigned to the $\hat{u}_{\ell}=1$ branch of Eq.~\ref{eq:directional_merge}, where $k$ matches the number of such layers in the corresponding \ours{} recipe. Because this layer pattern is fixed, the complement symmetry used by \ours{} does not make $\alpha$ and $1-\alpha$ redundant; complementing the pattern would generally also change its cardinality from $k$ to $L-k$. We therefore sweep the full grid $\alpha\in\{0.0,0.1,\ldots,1.0\}$ for \textsc{Random-k}. Table~\ref{tab:full_alpha_grid} reports the complete sweep.

\begin{table*}[t]
\caption{\textbf{Full alpha-grid results on the TP/TR diagnostic set.} Each entry reports the search objective $\mathcal{F}(g)$ for the corresponding interpolation coefficient. For each \textsc{Random-k} run, the binary layer pattern is fixed while $\alpha$ is swept across the full grid. Bold marks the selected $\alpha$; objective ties are broken in favor of the higher TP score to avoid unnecessary perceptual degradation for TR improvement.}
\vspace{0.3em}
\centering
\small
\renewcommand{\arraystretch}{1.08}
\setlength{\tabcolsep}{3.2pt}
\sisetup{detect-weight=true, detect-inline-weight=math}

\begin{adjustbox}{max width=\textwidth}
\begin{tabular}{c | l | c |
                *{11}{S[table-format=2.1]}}
\toprule
Model & Variant & Run
& {$0.0$} & {$0.1$} & {$0.2$} & {$0.3$} & {$0.4$} & {$0.5$}
& {$0.6$} & {$0.7$} & {$0.8$} & {$0.9$} & {$1.0$} \\
\midrule

\multirow[c]{7}{*}[-2ex]{LongVA-7B}
& \textsc{Full-backbone} & --
& \multicolumn{1}{c}{--}
& 38.4
& {\bfseries 40.1}
& 39.5
& 37.3
& 33.9
& 31.1
& 29.9
& 19.8
& 4.5
& \multicolumn{1}{c}{-1.1} \\
\cmidrule(lr){2-14}

& \textsc{All-attention} & --
& \multicolumn{1}{c}{--}
& {\bfseries 39.5}
& 39.0
& 39.5
& 39.0
& 39.5
& 37.9
& 39.0
& 37.3
& 35.6
& 32.8 \\
\cmidrule(lr){2-14}

& \multirow{5}{*}{\makecell[c]{\textsc{Random-k}\\\textsc{attention}}}
& 1
& 36.2
& 39.0
& 39.0
& 38.4
& 39.0
& 39.5
& 39.5
& 39.5
& 39.5
& 39.0
& {\bfseries 40.1} \\

& & 2
& 37.9
& 37.3
& 37.3
& 39.0
& 39.0
& 39.5
& 39.0
& 39.5
& 39.5
& {\bfseries 41.2}
& 38.4 \\

& & 3
& 38.4
& 39.0
& 39.5
& 39.0
& 39.5
& {\bfseries 39.5}
& 39.5
& 39.0
& 38.4
& 36.7
& 37.9 \\

& & 4
& 39.0
& 41.2
& {\bfseries 41.8}
& 40.1
& 39.5
& 39.5
& 39.0
& 37.9
& 36.2
& 36.2
& 35.6 \\

& & 5
& 35.0
& 36.2
& 35.6
& 37.9
& 37.3
& 39.5
& 39.5
& 40.1
& 40.1
& {\bfseries 40.7}
& 38.4 \\

\midrule

\multirow[c]{7}{*}[-2ex]{InternVL3-8B}
& \textsc{Full-backbone} & --
& \multicolumn{1}{c}{--}
& 53.1
& {\bfseries 53.1}
& 50.3
& 49.2
& 48.6
& 42.9
& 40.1
& 36.7
& 32.2
& 27.7 \\
\cmidrule(lr){2-14}

& \textsc{All-attention} & --
& \multicolumn{1}{c}{--}
& {\bfseries 54.2}
& 53.7
& 54.2
& 54.2
& 51.4
& 51.4
& 50.8
& 49.2
& 49.2
& 50.3 \\
\cmidrule(lr){2-14}

& \multirow{5}{*}{\makecell[c]{\textsc{Random-k}\\\textsc{attention}}}
& 1
& 52.0
& 52.0
& 51.4
& 50.8
& 51.4
& 51.4
& {\bfseries 52.0}
& 50.8
& 51.4
& 51.4
& 50.3 \\

& & 2
& 49.7
& 51.4
& 52.5
& {\bfseries 53.7}
& 52.5
& 51.4
& 52.0
& 52.5
& 51.4
& 50.8
& 50.3 \\

& & 3
& 52.5
& 51.4
& 52.5
& {\bfseries 52.5}
& 52.0
& 51.4
& 52.5
& 52.0
& 52.5
& 51.4
& 52.0 \\

& & 4
& 52.0
& 50.3
& 52.0
& 52.0
& 52.5
& 51.4
& {\bfseries 53.1}
& 52.5
& 52.0
& 51.4
& 49.2 \\

& & 5
& 48.0
& 49.7
& 51.4
& 51.4
& 52.5
& 51.4
& {\bfseries 54.2}
& 53.7
& 52.0
& 53.1
& 51.4 \\

\midrule

\multirow[c]{7}{*}[-2ex]{Qwen3-VL-4B}
& \textsc{Full-backbone} & --
& \multicolumn{1}{c}{--}
& 42.9
& 44.1
& {\bfseries 45.2}
& 44.1
& 39.5
& 35.6
& 32.2
& 26.0
& 24.3
& 13.6 \\
\cmidrule(lr){2-14}

& \textsc{All-attention} & --
& \multicolumn{1}{c}{--}
& {\bfseries 44.1}
& 42.9
& 42.9
& 43.5
& 41.2
& 39.5
& 40.1
& 39.0
& 35.0
& 33.3 \\
\cmidrule(lr){2-14}

& \multirow{5}{*}{\makecell[c]{\textsc{Random-k}\\\textsc{attention}}}
& 1
& 39.0
& 38.4
& 41.2
& 41.8
& 42.4
& 41.2
& {\bfseries 45.2}
& 44.6
& 43.5
& 42.9
& 42.9 \\

& & 2
& 40.7
& 42.4
& {\bfseries 43.5}
& 42.9
& 42.4
& 41.2
& 41.2
& 41.2
& 39.5
& 39.5
& 37.3 \\

& & 3
& 40.7
& 42.9
& {\bfseries 45.8}
& 44.6
& 42.4
& 41.2
& 41.2
& 41.8
& 41.8
& 42.4
& 38.4 \\

& & 4
& 43.5
& 43.5
& 43.5
& 42.9
& {\bfseries 44.1}
& 41.2
& 40.7
& 40.7
& 42.4
& 39.0
& 37.3 \\

& & 5
& 38.4
& 39.0
& 38.4
& 38.4
& 42.4
& 41.2
& {\bfseries 42.9}
& 42.4
& 41.2
& 37.9
& 42.4 \\

\bottomrule
\end{tabular}
\end{adjustbox}

\label{tab:full_alpha_grid}
\end{table*}

The preferred interpolation coefficient depends on both intervention granularity and layer identity. \textsc{Full-Backbone} generally favors small $\alpha$ and degrades as more weight is placed on the pre-adaptation checkpoint, while \textsc{All-Attention} selects $\alpha=0.1$ for all three models. For \textsc{Random-k}, the preferred coefficient varies substantially across fixed random layer patterns, showing that interpolation direction and layer identity are coupled.

\paragraph{Alpha-tuned \textsc{Random-k} controls for layer identity.}
Table~\ref{tab:appendix_randomk_layers} reports the five random layer patterns used for each backbone. Every run assigns exactly $k$ self-attention layers to the $\hat{u}_{\ell}=1$ branch, with $k$ matching the corresponding \ours{} recipe, while the remaining layers follow the $\hat{u}_{\ell}=0$ branch. For each fixed pattern, $\alpha$ is tuned independently using the grid in Table~\ref{tab:full_alpha_grid}. These configurations form the $\alpha$-tuned \textsc{Random-k} baseline reported in the main results.

\begin{table*}[t]
\caption{\textbf{Alpha-tuned random-$k$ baseline recipes.}
For each model, we report five independent random binary layer patterns, each assigning exactly $k$ self-attention layers to the $\hat{u}_{\ell}=1$ branch of Eq.~\ref{eq:directional_merge}, with $k$ matching the number of layers assigned $\hat{u}_{\ell}^{\ast}=1$ in the corresponding \ours{} recipe. The interpolation coefficient $\alpha$ is tuned independently for every run on the TP/TR diagnostic set. Because $\alpha$ may be below $0.5$, the $\hat{u}_{\ell}=1$ branch is not necessarily LLM-dominant. Layer indices are 0-based.}
\vspace{0.3em}
\centering
\footnotesize
\setlength{\tabcolsep}{5pt}
\renewcommand{\arraystretch}{1.05}

\begin{tabularx}{\textwidth}{
    l
    c
    c
    c
    >{\raggedright\arraybackslash}X
}
\toprule
\textbf{VLM}
& $\boldsymbol{\alpha}$
& \textbf{\#Layers}
& \textbf{Run}
& $\boldsymbol{\hat{u}_{\ell}=1}$ \textbf{ self-attention layer indices} \\
\midrule

\multirow[c]{5}{*}{LongVA-7B}
& 1.0
& \multirow[c]{5}{*}{11}
& 1
& 5, 7--8, 10--11, 17, 20--22, 26--27 \\

& 0.9
&
& 2
& 0, 3--6, 8--9, 14--15, 19, 27 \\

& 0.5
&
& 3
& 1, 4, 7, 11--12, 15, 20--21, 23, 26--27 \\

& 0.2
&
& 4
& 1, 6--7, 15, 17--18, 20--22, 25, 27 \\

& 0.9
&
& 5
& 5, 8--9, 12--15, 21, 24--25, 27 \\

\midrule

\multirow[c]{5}{*}{InternVL3-8B}
& 0.6
& \multirow[c]{5}{*}{9}
& 1
& 5, 8, 10, 12--13, 16, 23--24, 27 \\

& 0.3
&
& 2
& 0--1, 3--4, 6, 10, 12, 16, 23 \\

& 0.3
&
& 3
& 0, 2--3, 9, 13, 15--16, 18, 26 \\

& 0.6
&
& 4
& 0--1, 4, 12, 16, 19, 24, 26--27 \\

& 0.6
&
& 5
& 1, 6, 8, 10, 18--19, 22, 26--27 \\

\midrule

\multirow[c]{5}{*}{Qwen3-VL-4B}
& 0.6
& \multirow[c]{5}{*}{16}
& 1
& 2--3, 6--7, 12, 14, 21--25, 28--31, 33 \\

& 0.2
&
& 2
& 0, 3--7, 9--11, 15, 17, 22, 24, 30, 33, 35 \\

& 0.2
&
& 3
& 1--4, 7--8, 10, 14--15, 17--18, 20--23, 25 \\

& 0.4
&
& 4
& 0--2, 7, 11, 14--15, 18--19, 21, 25--28, 32, 35 \\

& 0.6
&
& 5
& 0, 3--4, 6, 8--10, 14, 17--18, 20--21, 24, 30--31, 35 \\

\bottomrule
\end{tabularx}

\label{tab:appendix_randomk_layers}
\end{table*}

Independent tuning prevents interpolation coefficient selection from favoring \ours{}. Since the random patterns match the cardinality of the \ours{} pattern but not its layer identities, their remaining performance gap supports the contribution of the discovered layer configuration.

\section{Benchmark Details and Complete Results}
\label{app:full_results}

This section summarizes the evaluated benchmarks and reports the complete per-benchmark results underlying the main-text comparisons.

\subsection{Benchmark Statistics and Task Formats}
\label{sec:benchmark_details}

\begin{itemize}[leftmargin=*]
    \item \textbf{Video-MME} \citep{fu2025video}: A broad VideoQA benchmark covering diverse video-understanding capabilities, including long-context temporal reasoning. It contains 900 videos and 2,700 multiple-choice questions.

    \item \textbf{LongVideoBench} \citep{wu2024longvideobench}: A long-context video--language benchmark requiring retrieval and reasoning over temporally localized segments. Its validation set contains 752 videos and 1,337 multiple-choice questions.

    \item \textbf{LVBench} \citep{wang2025lvbench}: A long-video understanding benchmark covering temporal grounding, reasoning, and related capabilities. It contains 103 videos longer than 30 minutes and 1,549 question--answer pairs with temporal annotations.

    \item \textbf{MMBench-Video} \citep{fang2024mmbench}: A VideoQA benchmark for fine-grained multi-shot understanding and temporal reasoning, with 609 videos and 1,998 open-ended question--answer pairs evaluated using an LLM-based scoring protocol.

    \item \textbf{Video-Holmes} \citep{cheng2025video}: A reasoning-centric benchmark built from suspense videos, designed to test multi-step inference through cross-segment clue aggregation. It contains 270 videos and 1,837 multiple-choice questions.
\end{itemize}

\subsection{Complete Results by Benchmark}
\label{sec:full_results}

We report the complete metrics underlying the summarized main-text results: Table~\ref{tab:video_mme_full} for Video-MME, Table~\ref{tab:longvideobench_full} for LongVideoBench, Table~\ref{tab:lvbench} for LVBench, Table~\ref{tab:mmbench_video} for MMBench-Video, and Table~\ref{tab:video_holmes_full} for Video-Holmes.

\begin{table*}[t]
\caption{\textbf{Full results (Accuracy \%) on Video-MME.}
TP: Temporal Perception;
SP: Spatial Perception;
AP: Attribute Perception;
ActRec: Action Recognition;
ObjRec: Object Recognition;
OCR: OCR Problems;
Cnt: Counting Problem;
TR: Temporal Reasoning;
SR: Spatial Reasoning;
AR: Action Reasoning;
OR: Object Reasoning;
IS: Information Synopsis.}
\vspace{0.3em}
\centering
\small
\renewcommand{\arraystretch}{1.08}
\setlength{\tabcolsep}{3.5pt}

\begin{adjustbox}{max width=\textwidth}
\begin{tabular}{c | l | c |
                S[table-format=2.1] |
                S[table-format=2.1]
                S[table-format=2.1]
                S[table-format=2.1]
                S[table-format=2.1]
                S[table-format=2.1]
                S[table-format=2.1]
                S[table-format=2.1]
                S[table-format=2.1]
                S[table-format=2.1]
                S[table-format=2.1]
                S[table-format=2.1]
                S[table-format=2.1]}
\toprule
Model & Variant & Run & {Overall} & {TP} & {SP} & {AP} & {ActRec} &
{ObjRec} & {OCR} & {Cnt} & {TR} & {SR} & {AR} & {OR} & {IS} \\
\midrule

\multirow[c]{16}{*}[-2ex]{LongVA-7B}
& \textsc{Base} & --
& 54.1 & 60.0 & 64.8 & 67.1 & 53.4 & 60.2 & 55.4 & 30.6 & 40.1 & 62.5 & 48.1 & 52.4 & 69.7 \\
\cmidrule(lr){2-16}

& \textsc{LoRA} & --
& 55.0 & 72.7 & 66.7 & 67.6 & 54.3 & 58.8 & 56.1 & 31.0 & 53.1 & 67.9 & 48.4 & 50.0 & 69.0 \\
\cmidrule(lr){2-16}

& \textsc{FFT} & --
& 54.3 & 81.8 & 63.0 & 65.8 & 50.8 & 54.8 & 55.4 & 32.1 & 64.4 & 67.9 & 44.6 & 50.2 & 67.5 \\
\cmidrule(lr){2-16}

& \textsc{Full-backbone} & --
& 54.1 & 60.0 & 63.0 & 66.2 & 51.8 & 59.3 & 55.4 & 31.0 & 40.1 & 66.1 & 49.1 & 52.4 & 70.9 \\
\cmidrule(lr){2-16}

& \textsc{All-attention} & --
& 54.3 & 61.8 & 64.8 & 65.8 & 53.4 & 59.3 & 56.1 & 31.0 & 39.5 & 67.9 & 48.8 & 52.6 & 70.6 \\
\cmidrule(lr){2-16}

& \multirow{5}{*}{\textsc{Random-k}} & 1
& 54.2 & 60.0 & 64.8 & 67.6 & 52.1 & 59.0 & 55.4 & 30.6 & 40.1 & 66.1 & 49.5 & 52.0 & 71.2 \\

& & 2
& 53.3 & 60.0 & 57.4 & 63.1 & 50.8 & 60.2 & 56.1 & 28.7 & 38.4 & 62.5 & 47.7 & 52.6 & 71.2 \\

& & 3
& 53.6 & 56.4 & 66.7 & 64.0 & 50.8 & 59.6 & 58.3 & 32.5 & 39.0 & 66.1 & 48.4 & 51.5 & 69.0 \\

& & 4
& 53.2 & 60.0 & 64.8 & 63.1 & 51.1 & 59.3 & 54.7 & 29.1 & 35.6 & 64.3 & 50.2 & 51.3 & 70.9 \\

& & 5
& 54.4 & 58.2 & 63.0 & 69.8 & 53.4 & 59.6 & 54.0 & 31.7 & 39.0 & 67.9 & 50.5 & 50.9 & 70.9 \\
\cmidrule(lr){2-16}

& \multirow{5}{*}{\shortstack{\textsc{Random-k}\\[-1pt]($\alpha$-tuned)}} & 1
& 54.2 & 60.0 & 64.8 & 67.6 & 52.1 & 59.0 & 55.4 & 30.6 & 40.1 & 66.1 & 49.5 & 52.0 & 71.2 \\

& & 2
& 53.8 & 60.0 & 59.3 & 64.0 & 50.2 & 61.0 & 56.8 & 31.0 & 41.2 & 64.3 & 47.4 & 52.2 & 71.2 \\

& & 3
& 53.7 & 60.0 & 63.0 & 65.3 & 50.5 & 60.5 & 55.4 & 29.5 & 39.5 & 64.3 & 48.8 & 52.4 & 70.3 \\

& & 4
& 53.4 & 60.0 & 59.3 & 65.8 & 49.5 & 60.7 & 56.8 & 26.1 & 41.8 & 69.6 & 48.1 & 51.5 & 70.9 \\

& & 5
& 54.7 & 58.2 & 63.0 & 69.8 & 53.7 & 58.8 & 55.4 & 31.3 & 41.2 & 67.9 & 50.2 & 51.8 & 70.9 \\
\cmidrule(lr){2-16}

& \ourcell  \ours{} & \ourcell  --
& \ourcell  55.1 & \ourcell  61.8 & \ourcell  64.8 & \ourcell  68.0 & \ourcell  52.7 & \ourcell  59.9 & \ourcell  57.6 & \ourcell  29.9 & \ourcell  49.7 & \ourcell  69.6 & \ourcell  48.1 & \ourcell  53.3 & \ourcell  69.3 \\

\midrule

\multirow[c]{16}{*}[-2ex]{InternVL3-8B}
& \textsc{Base} & --
& 66.5 & 74.5 & 64.8 & 78.8 & 64.2 & 72.6 & 72.7 & 46.6 & 52.0 & 83.9 & 61.4 & 64.1 & 79.3 \\
\cmidrule(lr){2-16}

& \textsc{LoRA} & --
& 68.0 & 83.6 & 64.8 & 78.8 & 65.8 & 72.3 & 71.9 & 45.9 & 71.2 & 82.1 & 61.1 & 63.9 & 80.2 \\
\cmidrule(lr){2-16}

& \textsc{FFT} & --
& 67.0 & 85.5 & 63.0 & 77.0 & 65.2 & 72.0 & 69.1 & 47.0 & 71.2 & 76.8 & 60.4 & 61.9 & 78.6 \\
\cmidrule(lr){2-16}

& \textsc{Full-backbone} & --
& 66.6 & 76.4 & 64.8 & 77.0 & 66.8 & 70.9 & 71.9 & 47.0 & 53.1 & 82.1 & 61.1 & 63.7 & 80.8 \\
\cmidrule(lr){2-16}

& \textsc{All-attention} & --
& 66.6 & 76.4 & 63.0 & 77.9 & 64.2 & 72.6 & 71.2 & 48.5 & 54.2 & 83.9 & 61.1 & 63.7 & 79.6 \\
\cmidrule(lr){2-16}

& \multirow{5}{*}{\textsc{Random-k}} & 1
& 66.6 & 80.0 & 66.7 & 77.9 & 66.1 & 72.9 & 71.9 & 46.6 & 51.4 & 83.9 & 58.9 & 63.7 & 80.5 \\

& & 2
& 67.0 & 78.2 & 64.8 & 76.1 & 66.1 & 73.4 & 76.3 & 50.4 & 50.8 & 80.4 & 60.0 & 64.1 & 79.9 \\

& & 3
& 66.4 & 76.4 & 68.5 & 78.4 & 63.9 & 71.5 & 72.7 & 47.4 & 51.4 & 83.9 & 60.7 & 63.9 & 79.9 \\

& & 4
& 66.6 & 81.8 & 63.0 & 77.5 & 64.5 & 72.6 & 73.4 & 46.6 & 51.4 & 85.7 & 60.0 & 64.5 & 79.9 \\

& & 5
& 66.6 & 74.5 & 68.5 & 76.1 & 65.2 & 73.2 & 73.4 & 47.0 & 53.1 & 87.5 & 60.7 & 62.6 & 80.5 \\
\cmidrule(lr){2-16}

& \multirow{5}{*}{\shortstack{\textsc{Random-k}\\[-1pt]($\alpha$-tuned)}} & 1
& 66.7 & 80.0 & 68.5 & 77.0 & 66.1 & 72.0 & 71.2 & 50.4 & 52.0 & 83.9 & 59.6 & 63.2 & 79.6 \\

& & 2
& 66.5 & 76.4 & 66.7 & 77.9 & 65.5 & 72.6 & 71.9 & 47.4 & 53.7 & 83.9 & 60.7 & 63.0 & 78.9 \\

& & 3
& 66.2 & 81.8 & 64.8 & 75.7 & 66.5 & 71.2 & 71.2 & 48.1 & 52.5 & 85.7 & 59.3 & 62.6 & 79.9 \\

& & 4
& 66.5 & 78.2 & 68.5 & 77.0 & 66.5 & 72.0 & 71.9 & 47.4 & 53.1 & 87.5 & 60.0 & 62.8 & 79.3 \\

& & 5
& 66.8 & 76.4 & 68.5 & 76.6 & 66.8 & 72.3 & 72.7 & 48.9 & 54.2 & 89.3 & 60.7 & 62.3 & 79.3 \\
\cmidrule(lr){2-16}

 & \ourcell  \ours{} & \ourcell  --
& \ourcell  67.5 & \ourcell  81.8 & \ourcell  63.0 & \ourcell  77.5 & \ourcell  67.7 & \ourcell  71.5 & \ourcell  75.5 & \ourcell  45.9 & \ourcell  57.6 & \ourcell  82.1 & \ourcell  62.5 & \ourcell  64.5 & \ourcell  80.5 \\

\midrule

\multirow[c]{16}{*}[-2ex]{Qwen3-VL-4B}
& \textsc{Base} & --
& 64.0 & 69.1 & 72.2 & 76.6 & 62.6 & 71.8 & 74.1 & 47.8 & 45.2 & 75.0 & 55.8 & 59.5 & 77.4 \\
\cmidrule(lr){2-16}

& \textsc{LoRA} & --
& 64.7 & 76.4 & 74.1 & 76.1 & 64.5 & 72.0 & 74.1 & 46.3 & 57.1 & 76.8 & 54.7 & 60.8 & 73.1 \\
\cmidrule(lr){2-16}

& \textsc{FFT} & --
& 65.0 & 85.5 & 75.9 & 74.8 & 67.4 & 69.5 & 71.9 & 41.4 & 70.6 & 87.5 & 54.7 & 58.1 & 74.0 \\
\cmidrule(lr){2-16}

& \textsc{Full-backbone} & --
& 63.5 & 67.3 & 70.4 & 78.8 & 62.0 & 71.2 & 71.9 & 44.0 & 45.8 & 75.0 & 57.5 & 57.7 & 77.7 \\
\cmidrule(lr){2-16}

& \textsc{All-attention} & --
& 64.1 & 67.3 & 72.2 & 76.6 & 63.3 & 72.3 & 73.4 & 48.9 & 44.1 & 75.0 & 55.4 & 59.5 & 77.4 \\
\cmidrule(lr){2-16}

& \multirow{5}{*}{\textsc{Random-k}} & 1
& 63.1 & 65.5 & 72.2 & 78.8 & 61.3 & 71.2 & 69.8 & 47.0 & 44.1 & 76.8 & 55.4 & 57.7 & 76.5 \\

& & 2
& 60.4 & 60.0 & 64.8 & 76.1 & 59.4 & 67.5 & 71.9 & 41.0 & 40.1 & 75.0 & 53.0 & 56.4 & 73.7 \\

& & 3
& 60.0 & 67.3 & 63.0 & 75.2 & 60.7 & 68.9 & 71.2 & 36.9 & 39.0 & 76.8 & 51.6 & 55.7 & 73.7 \\

& & 4
& 61.2 & 61.8 & 72.2 & 75.2 & 61.3 & 67.2 & 69.1 & 43.3 & 39.5 & 73.2 & 52.6 & 58.1 & 76.2 \\

& & 5
& 61.8 & 61.8 & 63.0 & 77.0 & 60.1 & 68.9 & 70.5 & 40.3 & 44.6 & 75.0 & 54.7 & 58.1 & 77.7 \\
\cmidrule(lr){2-16}

& \multirow{5}{*}{\shortstack{\textsc{Random-k}\\[-1pt]($\alpha$-tuned)}} & 1
& 59.0 & 70.9 & 66.7 & 71.6 & 57.2 & 64.4 & 65.5 & 41.8 & 45.2 & 67.9 & 51.2 & 57.5 & 69.3 \\

& & 2
& 58.5 & 67.3 & 63.0 & 71.6 & 56.9 & 63.8 & 64.7 & 37.3 & 44.1 & 67.9 & 51.6 & 59.0 & 69.3 \\

& & 3
& 63.1 & 69.1 & 72.2 & 81.1 & 63.6 & 69.2 & 72.7 & 42.2 & 45.8 & 78.6 & 53.3 & 58.4 & 76.5 \\

& & 4
& 63.3 & 70.9 & 72.2 & 78.4 & 62.6 & 71.2 & 73.4 & 44.4 & 44.1 & 76.8 & 54.0 & 59.0 & 75.9 \\

& & 5
& 62.8 & 69.1 & 72.2 & 79.3 & 61.0 & 70.1 & 73.4 & 41.4 & 42.9 & 76.8 & 55.1 & 59.0 & 76.2 \\
\cmidrule(lr){2-16}

 & \ourcell  \ours{} & \ourcell  --
& \ourcell  63.6 & \ourcell  70.9 & \ourcell  72.2 & \ourcell  78.8 & \ourcell  62.3 & \ourcell  71.5 & \ourcell  71.2 & \ourcell  43.3 & \ourcell  46.9 & \ourcell  75.0 & \ourcell  55.8 & \ourcell  58.8 & \ourcell  77.1 \\

\bottomrule
\end{tabular}
\end{adjustbox}
\label{tab:video_mme_full}
\end{table*}

\begin{table*}[t]
\caption{\textbf{Full results (Accuracy \%) on LongVideoBench.}
The columns S2E, S2O, S2A, E2O, O2E, T2E, T2O, and T2A belong to the
higher-level category \textit{Perception}, while E3E, O3O, SSS, SOS, SAA,
T3E, T3O, TOS, and TAA belong to \textit{Relation}.}
\vspace{0.3em}
\centering
\scriptsize
\renewcommand{\arraystretch}{1.08}
\setlength{\tabcolsep}{3pt}

\begin{adjustbox}{max width=\textwidth}
\begin{tabular}{c | l | c |
                S[table-format=2.1] |
                S[table-format=2.1]
                S[table-format=2.1]
                S[table-format=2.1]
                S[table-format=2.1]
                S[table-format=2.1]
                S[table-format=2.1]
                S[table-format=2.1]
                S[table-format=2.1] |
                S[table-format=2.1]
                S[table-format=2.1]
                S[table-format=2.1]
                S[table-format=2.1]
                S[table-format=2.1]
                S[table-format=2.1]
                S[table-format=2.1]
                S[table-format=2.1]
                S[table-format=2.1]}
\toprule
\multirow{2}{*}{Model}
& \multirow{2}{*}{Variant}
& \multirow{2}{*}{Run}
& \multicolumn{1}{c|}{\multirow{2}{*}{Overall}}
& \multicolumn{8}{c|}{Perception}
& \multicolumn{9}{c}{Relation} \\
\cmidrule(lr){5-12}
\cmidrule(lr){13-21}
& & &
& {S2E} & {S2O} & {S2A} & {E2O} & {O2E} & {T2E} & {T2O} & {T2A}
& {E3E} & {O3O} & {SSS} & {SOS} & {SAA} & {T3E} & {T3O} & {TOS} & {TAA} \\
\midrule

\multirow[c]{16}{*}[-2ex]{LongVA-7B}
& \textsc{Base} & --
& 52.5 & 60.2 & 48.6 & 75.0 & 52.3 & 58.6 & 55.4 & 51.3 & 54.4
& 56.4 & 51.5 & 35.1 & 65.4 & 51.4 & 43.8 & 50.0 & 35.6 & 43.9 \\
\cmidrule(lr){2-21}

& \textsc{LoRA} & --
& 52.2 & 60.2 & 47.2 & 73.9 & 52.3 & 60.9 & 56.9 & 52.6 & 54.4
& 54.3 & 48.5 & 34.0 & 64.2 & 50.0 & 45.2 & 52.7 & 31.5 & 45.1 \\
\cmidrule(lr){2-21}

& \textsc{FFT} & --
& 40.8 & 45.2 & 36.1 & 67.0 & 44.6 & 49.4 & 50.8 & 38.2 & 40.5
& 45.7 & 40.9 & 21.6 & 38.3 & 37.5 & 34.2 & 43.2 & 27.4 & 32.9 \\
\cmidrule(lr){2-21}

& \textsc{Full-backbone} & --
& 52.9 & 62.4 & 52.8 & 76.1 & 50.8 & 59.8 & 58.5 & 51.3 & 51.9
& 55.3 & 48.5 & 34.0 & 66.7 & 54.2 & 45.2 & 50.0 & 35.6 & 42.7 \\
\cmidrule(lr){2-21}

& \textsc{All-attention} & --
& 52.6 & 60.2 & 48.6 & 75.0 & 50.8 & 58.6 & 55.4 & 52.6 & 55.7
& 56.4 & 53.0 & 33.0 & 66.7 & 52.8 & 43.8 & 50.0 & 34.2 & 43.9 \\
\cmidrule(lr){2-21}

& \multirow{5}{*}{\textsc{Random-k}} & 1
& 52.8 & 62.4 & 50.0 & 75.0 & 46.2 & 57.5 & 53.8 & 50.0 & 53.2
& 59.6 & 50.0 & 36.1 & 65.4 & 54.2 & 46.6 & 50.0 & 37.0 & 45.1 \\

& & 2
& 52.5 & 59.1 & 56.9 & 75.0 & 58.5 & 59.8 & 55.4 & 51.3 & 48.1
& 50.0 & 45.5 & 33.0 & 67.9 & 52.8 & 45.2 & 54.1 & 34.2 & 45.1 \\

& & 3
& 52.8 & 60.2 & 51.4 & 75.0 & 50.8 & 59.8 & 60.0 & 53.9 & 51.9
& 55.3 & 51.5 & 33.0 & 65.4 & 50.0 & 47.9 & 51.4 & 31.5 & 46.3 \\

& & 4
& 52.6 & 60.2 & 52.8 & 76.1 & 49.2 & 58.6 & 60.0 & 50.0 & 53.2
& 53.2 & 50.0 & 32.0 & 65.4 & 52.8 & 46.6 & 50.0 & 37.0 & 45.1 \\

& & 5
& 53.0 & 60.2 & 52.8 & 76.1 & 49.2 & 60.9 & 55.4 & 53.9 & 53.2
& 55.3 & 53.0 & 30.9 & 65.4 & 54.2 & 45.2 & 54.1 & 35.6 & 43.9 \\
\cmidrule(lr){2-21}

& \multirow{5}{*}{\shortstack{\textsc{Random-k}\\[-1pt]($\alpha$-tuned)}} & 1
& 52.8 & 62.4 & 50.0 & 75.0 & 46.2 & 57.5 & 53.8 & 50.0 & 53.2
& 59.6 & 50.0 & 36.1 & 65.4 & 54.2 & 46.6 & 50.0 & 37.0 & 45.1 \\

& & 2
& 52.7 & 58.1 & 56.9 & 76.1 & 58.5 & 58.6 & 56.9 & 52.6 & 48.1
& 51.1 & 45.5 & 30.9 & 65.4 & 52.8 & 45.2 & 55.4 & 37.0 & 46.3 \\

& & 3
& 52.7 & 58.1 & 55.6 & 75.0 & 52.3 & 55.2 & 60.0 & 52.6 & 51.9
& 54.3 & 48.5 & 34.0 & 66.7 & 54.2 & 43.8 & 51.4 & 34.2 & 47.6 \\

& & 4
& 52.8 & 58.1 & 59.7 & 73.9 & 55.4 & 57.5 & 60.0 & 51.3 & 50.6
& 54.3 & 48.5 & 37.1 & 63.0 & 51.4 & 43.8 & 52.7 & 32.9 & 46.3 \\

& & 5
& 53.3 & 60.2 & 55.6 & 75.0 & 47.7 & 60.9 & 56.9 & 56.6 & 51.9
& 55.3 & 53.0 & 30.9 & 65.4 & 55.6 & 45.2 & 55.4 & 35.6 & 42.7 \\
\cmidrule(lr){2-21}

& \ourcell  \ours{} & \ourcell  --
& \ourcell  53.6 & \ourcell  62.4 & \ourcell  55.6 & \ourcell  73.9 & \ourcell  50.8 & \ourcell  58.6 & \ourcell  58.5 & \ourcell  55.3 & \ourcell  55.7
& \ourcell  56.4 & \ourcell  48.5 & \ourcell  32.0 & \ourcell  65.4 & \ourcell  51.4 & \ourcell  43.8 & \ourcell  56.8 & \ourcell  38.4 & \ourcell  46.3 \\

\midrule

\multirow[c]{16}{*}[-2ex]{InternVL3-8B}
& \textsc{Base} & --
& 59.8 & 68.8 & 65.3 & 78.4 & 73.8 & 58.6 & 56.9 & 57.9 & 63.3
& 70.2 & 57.6 & 43.3 & 63.0 & 54.2 & 58.9 & 52.7 & 45.2 & 47.6 \\
\cmidrule(lr){2-21}

& \textsc{LoRA} & --
& 58.9 & 66.7 & 65.3 & 77.3 & 70.8 & 54.0 & 55.4 & 59.2 & 63.3
& 66.0 & 56.1 & 46.4 & 61.7 & 54.2 & 57.5 & 51.4 & 43.8 & 51.2 \\
\cmidrule(lr){2-21}

& \textsc{FFT} & --
& 55.7 & 64.5 & 55.6 & 75.0 & 63.1 & 51.7 & 50.8 & 59.2 & 54.4
& 61.7 & 50.0 & 43.3 & 61.7 & 51.4 & 61.6 & 48.6 & 38.4 & 52.4 \\
\cmidrule(lr){2-21}

& \textsc{Full-backbone} & --
& 59.9 & 68.8 & 62.5 & 78.4 & 73.8 & 57.5 & 60.0 & 57.9 & 63.3
& 71.3 & 57.6 & 44.3 & 64.2 & 50.0 & 57.5 & 55.4 & 42.5 & 51.2 \\
\cmidrule(lr){2-21}

& \textsc{All-attention} & --
& 59.5 & 64.5 & 63.9 & 77.3 & 72.3 & 57.5 & 58.5 & 60.5 & 64.6
& 71.3 & 59.1 & 45.4 & 63.0 & 54.2 & 57.5 & 51.4 & 43.8 & 46.3 \\
\cmidrule(lr){2-21}

& \multirow{5}{*}{\textsc{Random-k}} & 1
& 59.5 & 66.7 & 65.3 & 73.9 & 69.2 & 56.3 & 60.0 & 60.5 & 62.0
& 69.1 & 56.1 & 45.4 & 66.7 & 54.2 & 58.9 & 55.4 & 42.5 & 48.8 \\

& & 2
& 59.8 & 68.8 & 62.5 & 75.0 & 72.3 & 57.5 & 60.0 & 57.9 & 62.0
& 70.2 & 57.6 & 42.3 & 67.9 & 52.8 & 60.3 & 55.4 & 42.5 & 51.2 \\

& & 3
& 58.7 & 68.8 & 62.5 & 78.4 & 66.2 & 58.6 & 60.0 & 56.6 & 60.8
& 69.1 & 63.6 & 42.3 & 59.3 & 54.2 & 53.4 & 52.7 & 41.1 & 48.8 \\

& & 4
& 60.1 & 65.6 & 65.3 & 73.9 & 73.8 & 56.3 & 58.5 & 60.5 & 64.6
& 70.2 & 59.1 & 45.4 & 66.7 & 51.4 & 60.3 & 55.4 & 45.2 & 48.8 \\

& & 5
& 59.3 & 65.6 & 62.5 & 76.1 & 72.3 & 56.3 & 61.5 & 57.9 & 62.0
& 71.3 & 54.5 & 44.3 & 69.1 & 54.2 & 56.2 & 51.4 & 42.5 & 48.8 \\
\cmidrule(lr){2-21}

& \multirow{5}{*}{\shortstack{\textsc{Random-k}\\[-1pt]($\alpha$-tuned)}} & 1
& 59.8 & 67.7 & 63.9 & 79.5 & 69.2 & 57.5 & 61.5 & 60.5 & 64.6
& 67.0 & 56.1 & 42.3 & 67.9 & 56.9 & 57.5 & 54.1 & 41.1 & 48.8 \\

& & 2
& 59.9 & 71.0 & 66.7 & 76.1 & 69.2 & 57.5 & 58.5 & 59.2 & 65.8
& 69.1 & 56.1 & 43.3 & 66.7 & 58.3 & 58.9 & 51.4 & 39.7 & 48.8 \\

& & 3
& 60.0 & 66.7 & 63.9 & 76.1 & 70.8 & 57.5 & 63.1 & 61.8 & 62.0
& 67.0 & 56.1 & 43.3 & 69.1 & 55.6 & 58.9 & 54.1 & 42.5 & 51.2 \\

& & 4
& 60.1 & 65.6 & 63.9 & 75.0 & 70.8 & 58.6 & 61.5 & 59.2 & 64.6
& 69.1 & 56.1 & 43.3 & 70.4 & 55.6 & 57.5 & 52.7 & 43.8 & 52.4 \\

& & 5
& 60.4 & 68.8 & 65.3 & 77.3 & 70.8 & 56.3 & 61.5 & 61.8 & 64.6
& 69.1 & 56.1 & 44.3 & 70.4 & 56.9 & 57.5 & 52.7 & 43.8 & 48.8 \\
\cmidrule(lr){2-21}

& \ourcell  \ours{} & \ourcell  --
& \ourcell  60.3 & \ourcell  69.9 & \ourcell  61.1 & \ourcell  75.0 & \ourcell  69.2 & \ourcell  59.8 & \ourcell  63.1 & \ourcell  61.8 & \ourcell  67.1
& \ourcell  68.1 & \ourcell  54.5 & \ourcell  43.3 & \ourcell  65.4 & \ourcell  54.2 & \ourcell  61.6 & \ourcell  54.1 & \ourcell  39.7 & \ourcell  54.9 \\

\midrule

\multirow[c]{16}{*}[-2ex]{Qwen3-VL-4B}
& \textsc{Base} & --
& 61.9 & 66.7 & 62.5 & 81.8 & 63.1 & 70.1 & 64.6 & 61.8 & 74.7
& 60.6 & 63.6 & 42.3 & 72.8 & 63.9 & 56.2 & 51.4 & 41.1 & 54.9 \\
\cmidrule(lr){2-21}

& \textsc{LoRA} & --
& 61.1 & 66.7 & 63.9 & 79.5 & 64.6 & 69.0 & 64.6 & 57.9 & 70.9
& 58.5 & 62.1 & 48.5 & 67.9 & 61.1 & 57.5 & 56.8 & 39.7 & 48.8 \\
\cmidrule(lr){2-21}

& \textsc{FFT} & --
& 55.3 & 63.4 & 50.0 & 73.9 & 61.5 & 55.2 & 50.8 & 52.6 & 65.8
& 56.4 & 56.1 & 47.4 & 58.0 & 50.0 & 56.2 & 52.7 & 37.0 & 50.0 \\
\cmidrule(lr){2-21}

& \textsc{Full-backbone} & --
& 60.4 & 63.4 & 61.1 & 76.1 & 66.2 & 65.5 & 60.0 & 57.9 & 70.9
& 68.1 & 65.2 & 46.4 & 70.4 & 55.6 & 57.5 & 50.0 & 37.0 & 52.4 \\
\cmidrule(lr){2-21}

& \textsc{All-attention} & --
& 62.0 & 64.5 & 65.3 & 80.7 & 63.1 & 69.0 & 64.6 & 63.2 & 74.7
& 61.7 & 63.6 & 42.3 & 74.1 & 62.5 & 57.5 & 54.1 & 41.1 & 52.4 \\
\cmidrule(lr){2-21}

& \multirow{5}{*}{\textsc{Random-k}} & 1
& 60.2 & 66.7 & 61.1 & 78.4 & 63.1 & 66.7 & 64.6 & 60.5 & 70.9
& 66.0 & 65.2 & 42.3 & 71.6 & 54.2 & 56.2 & 50.0 & 34.2 & 50.0 \\

& & 2
& 59.8 & 66.7 & 54.2 & 77.3 & 70.8 & 64.4 & 60.0 & 59.2 & 68.4
& 62.8 & 60.6 & 42.3 & 72.8 & 58.3 & 57.5 & 48.6 & 39.7 & 52.4 \\

& & 3
& 59.2 & 62.4 & 62.5 & 77.3 & 67.7 & 58.6 & 60.0 & 57.9 & 65.8
& 64.9 & 63.6 & 38.1 & 70.4 & 59.7 & 57.5 & 51.4 & 38.4 & 52.4 \\

& & 4
& 60.0 & 62.4 & 59.7 & 76.1 & 60.0 & 63.2 & 61.5 & 59.2 & 65.8
& 64.9 & 63.6 & 46.4 & 72.8 & 59.7 & 57.5 & 51.4 & 41.1 & 52.4 \\

& & 5
& 59.8 & 64.5 & 61.1 & 75.0 & 67.7 & 63.2 & 61.5 & 59.2 & 68.4
& 64.9 & 57.6 & 39.2 & 75.3 & 61.1 & 54.8 & 48.6 & 41.1 & 53.7 \\
\cmidrule(lr){2-21}

& \multirow{5}{*}{\shortstack{\textsc{Random-k}\\[-1pt]($\alpha$-tuned)}} & 1
& 60.7 & 66.7 & 65.3 & 78.4 & 61.5 & 62.1 & 60.0 & 61.8 & 67.1
& 67.0 & 68.2 & 42.3 & 76.5 & 58.3 & 54.8 & 54.1 & 35.6 & 51.2 \\

& & 2
& 59.9 & 64.5 & 63.9 & 75.0 & 64.6 & 63.2 & 58.5 & 57.9 & 68.4
& 67.0 & 59.1 & 44.3 & 72.8 & 59.7 & 52.1 & 55.4 & 37.0 & 52.4 \\

& & 3
& 60.7 & 64.5 & 62.5 & 78.4 & 61.5 & 66.7 & 61.5 & 60.5 & 68.4
& 62.8 & 59.1 & 42.3 & 77.8 & 63.9 & 54.8 & 52.7 & 41.1 & 52.4 \\

& & 4
& 60.7 & 65.6 & 65.3 & 77.3 & 63.1 & 62.1 & 61.5 & 63.2 & 64.6
& 63.8 & 63.6 & 44.3 & 75.3 & 59.7 & 54.8 & 54.1 & 41.1 & 52.4 \\

& & 5
& 60.8 & 66.7 & 65.3 & 79.5 & 61.5 & 62.1 & 61.5 & 59.2 & 69.6
& 67.0 & 63.6 & 40.2 & 75.3 & 61.1 & 56.2 & 52.7 & 37.0 & 53.7 \\
\cmidrule(lr){2-21}

& \ourcell  \ours{} & \ourcell  --
& \ourcell  60.6 & \ourcell  64.5 & \ourcell  68.1 & \ourcell  78.4 & \ourcell  64.6 & \ourcell  62.1 & \ourcell  56.9 & \ourcell  53.9 & \ourcell  67.1
& \ourcell  61.7 & \ourcell  62.1 & \ourcell  46.4 & \ourcell  74.1 & \ourcell  62.5 & \ourcell  56.2 & \ourcell  58.1 & \ourcell  37.0 & \ourcell  54.9 \\

\bottomrule
\end{tabular}
\end{adjustbox}

\label{tab:longvideobench_full}
\end{table*}

\begin{table*}[t]
\caption{\textbf{Full results (Accuracy \%) on LVBench.}
Overall: Overall performance;
ER: Entity Recognition;
EU: Event Understanding;
KIR: Key Information Retrieval;
TG: Temporal Grounding;
Sum: Summarization;
Rea: Reasoning.}
\vspace{0.3em}
\centering
\small
\renewcommand{\arraystretch}{1.08}
\setlength{\tabcolsep}{5pt}

\begin{adjustbox}{max width=0.85\textwidth}
\begin{tabular}{c | l | c |
                S[table-format=2.1] |
                S[table-format=2.1]
                S[table-format=2.1]
                S[table-format=2.1]
                S[table-format=2.1]
                S[table-format=2.1]
                S[table-format=2.1]}
\toprule
Model & Variant & Run & {Overall} & {ER} & {EU} & {KIR} & {TG} & {Sum} & {Rea} \\
\midrule

\multirow[c]{16}{*}[-2ex]{LongVA-7B}
& \textsc{Base} & --
& 37.6 & 38.3 & 35.1 & 37.1 & 29.5 & 39.7 & 45.3 \\
\cmidrule(lr){2-10}

& \textsc{LoRA} & --
& 37.7 & 39.3 & 35.4 & 35.4 & 30.9 & 34.5 & 45.3 \\
\cmidrule(lr){2-10}

& \textsc{FFT} & --
& 35.4 & 36.5 & 32.8 & 35.7 & 31.4 & 24.1 & 41.8 \\
\cmidrule(lr){2-10}

& \textsc{Full-backbone} & --
& 38.1 & 38.4 & 36.2 & 37.8 & 32.3 & 37.9 & 44.8 \\
\cmidrule(lr){2-10}

& \textsc{All-attention} & --
& 38.0 & 39.0 & 35.5 & 36.4 & 31.4 & 41.4 & 46.3 \\
\cmidrule(lr){2-10}

& \multirow{5}{*}{\textsc{Random-k}} & 1
& 38.6 & 39.4 & 36.2 & 37.5 & 29.5 & 37.9 & 46.8 \\

& & 2
& 37.1 & 37.5 & 35.9 & 35.7 & 30.5 & 39.7 & 42.3 \\

& & 3
& 37.7 & 39.7 & 36.0 & 37.1 & 30.9 & 34.5 & 44.3 \\

& & 4
& 37.4 & 38.7 & 35.1 & 36.8 & 32.3 & 41.4 & 44.8 \\

& & 5
& 38.0 & 39.0 & 35.2 & 36.8 & 30.9 & 41.4 & 43.8 \\
\cmidrule(lr){2-10}

& \multirow{5}{*}{\shortstack{\textsc{Random-k}\\[-1pt]($\alpha$-tuned)}} & 1
& 38.6 & 39.4 & 36.2 & 37.5 & 29.5 & 37.9 & 46.8 \\

& & 2
& 36.7 & 37.8 & 34.8 & 34.7 & 30.5 & 43.1 & 42.3 \\

& & 3
& 37.1 & 38.4 & 35.5 & 36.8 & 31.8 & 39.7 & 44.8 \\

& & 4
& 38.0 & 39.1 & 37.1 & 35.4 & 31.8 & 39.7 & 46.3 \\

& & 5
& 38.3 & 39.1 & 35.4 & 38.1 & 32.3 & 41.4 & 46.3 \\
\cmidrule(lr){2-10}

& \ourcell  \ours{} & \ourcell  --
& \ourcell  39.7 & \ourcell  41.1 & \ourcell  37.4 & \ourcell  37.5 & \ourcell  33.2 & \ourcell  43.1 & \ourcell  46.8 \\

\midrule

\multirow[c]{16}{*}[-2ex]{InternVL3-8B}
& \textsc{Base} & --
& 44.8 & 45.1 & 43.0 & 47.8 & 37.3 & 29.3 & 51.2 \\
\cmidrule(lr){2-10}

& \textsc{LoRA} & --
& 44.7 & 44.3 & 44.5 & 47.8 & 35.0 & 27.6 & 52.2 \\
\cmidrule(lr){2-10}

& \textsc{FFT} & --
& 44.4 & 44.6 & 43.9 & 45.4 & 37.3 & 22.4 & 50.2 \\
\cmidrule(lr){2-10}

& \textsc{Full-backbone} & --
& 44.7 & 44.3 & 45.0 & 47.1 & 34.1 & 32.8 & 50.7 \\
\cmidrule(lr){2-10}

& \textsc{All-attention} & --
& 44.7 & 44.6 & 43.3 & 48.1 & 36.4 & 31.0 & 50.2 \\
\cmidrule(lr){2-10}

& \multirow{5}{*}{\textsc{Random-k}} & 1
& 44.3 & 44.2 & 43.7 & 46.4 & 36.4 & 31.0 & 48.3 \\

& & 2
& 43.8 & 43.0 & 43.9 & 45.4 & 35.9 & 32.8 & 50.2 \\

& & 3
& 44.5 & 43.4 & 44.0 & 47.8 & 35.9 & 36.2 & 50.2 \\

& & 4
& 45.0 & 45.3 & 43.9 & 47.8 & 36.8 & 31.0 & 51.7 \\

& & 5
& 43.2 & 42.8 & 42.3 & 47.1 & 35.5 & 27.6 & 48.8 \\
\cmidrule(lr){2-10}

& \multirow{5}{*}{\shortstack{\textsc{Random-k}\\[-1pt]($\alpha$-tuned)}} & 1
& 44.8 & 44.6 & 44.2 & 48.1 & 38.6 & 32.8 & 48.8 \\

& & 2
& 44.2 & 44.5 & 43.6 & 46.7 & 37.7 & 27.6 & 48.3 \\

& & 3
& 44.0 & 45.1 & 42.8 & 47.1 & 37.3 & 25.9 & 48.3 \\

& & 4
& 44.6 & 44.9 & 43.6 & 48.1 & 35.5 & 31.0 & 48.3 \\

& & 5
& 44.1 & 44.6 & 42.8 & 47.4 & 35.5 & 29.3 & 46.8 \\
\cmidrule(lr){2-10}

& \ourcell  \ours{} & \ourcell  --
& \ourcell  45.0 & \ourcell  44.8 & \ourcell  44.2 & \ourcell  47.4 & \ourcell  35.5 & \ourcell  29.3 & \ourcell  53.2 \\

\midrule

\multirow[c]{16}{*}[-2ex]{Qwen3-VL-4B}
& \textsc{Base} & --
& 39.4 & 39.6 & 39.6 & 39.5 & 30.9 & 29.3 & 39.3 \\
\cmidrule(lr){2-10}

& \textsc{LoRA} & --
& 40.4 & 39.9 & 39.7 & 39.9 & 36.4 & 29.3 & 42.3 \\
\cmidrule(lr){2-10}

& \textsc{FFT} & --
& 40.3 & 42.5 & 39.4 & 41.2 & 30.9 & 20.7 & 43.3 \\
\cmidrule(lr){2-10}

& \textsc{Full-backbone} & --
& 34.9 & 35.2 & 34.9 & 32.6 & 30.0 & 27.6 & 33.3 \\
\cmidrule(lr){2-10}

& \textsc{All-attention} & --
& 43.7 & 43.6 & 43.9 & 45.4 & 35.9 & 32.8 & 43.3 \\
\cmidrule(lr){2-10}

& \multirow{5}{*}{\textsc{Random-k}} & 1
& 41.6 & 43.0 & 39.6 & 43.3 & 34.1 & 24.1 & 44.3 \\

& & 2
& 38.3 & 39.4 & 35.9 & 37.8 & 34.1 & 19.0 & 40.8 \\

& & 3
& 38.3 & 39.7 & 36.5 & 41.9 & 34.5 & 20.7 & 38.8 \\

& & 4
& 39.1 & 38.1 & 38.2 & 40.9 & 38.2 & 34.5 & 41.8 \\

& & 5
& 37.9 & 39.3 & 36.9 & 37.5 & 30.9 & 22.4 & 36.8 \\
\cmidrule(lr){2-10}

& \multirow{5}{*}{\shortstack{\textsc{Random-k}\\[-1pt]($\alpha$-tuned)}} & 1
& 38.3 & 39.3 & 37.1 & 42.3 & 31.4 & 20.7 & 40.8 \\

& & 2
& 40.9 & 41.8 & 39.1 & 43.6 & 32.7 & 27.6 & 42.8 \\

& & 3
& 40.2 & 40.0 & 40.2 & 43.0 & 33.2 & 24.1 & 40.3 \\

& & 4
& 40.7 & 42.4 & 38.8 & 43.0 & 33.6 & 20.7 & 43.3 \\

& & 5
& 40.0 & 41.1 & 37.9 & 42.3 & 31.4 & 22.4 & 40.3 \\
\cmidrule(lr){2-10}

& \ourcell  \ours{} & \ourcell  --
& \ourcell  40.6 & \ourcell  40.6 & \ourcell  39.9 & \ourcell  41.2 & \ourcell  32.3 & \ourcell  25.9 & \ourcell  46.3 \\

\bottomrule
\end{tabular}
\end{adjustbox}
\label{tab:lvbench}
\end{table*}

\begin{table*}[t]
\caption{\textbf{Full results on MMBench-Video.}
CP: Coarse Perception;
FP-S: Fine-grained Perception (Single-Instance);
FP-C: Fine-grained Perception (Cross-Instance);
HL: Hallucination;
LR: Logic Reasoning;
AR: Attribute Reasoning;
RR: Relation Reasoning;
CSR: Common Sense Reasoning;
TR: Temporal Reasoning.
All scores are based on a 3-grade marking scheme: 0 for worst, 3 for best.}
\vspace{0.3em}
\centering
\small
\renewcommand{\arraystretch}{1.08}
\setlength{\tabcolsep}{4pt}

\begin{adjustbox}{max width=0.9\textwidth}
\begin{tabular}{c | l | c |
                S[table-format=1.2] |
                S[table-format=1.2]
                S[table-format=1.2]
                S[table-format=1.2]
                S[table-format=1.2]
                S[table-format=1.2]
                S[table-format=1.2]
                S[table-format=1.2]
                S[table-format=1.2]
                S[table-format=1.2]}
\toprule
Model & Variant & Run & {Overall} & {CP} & {FP-S} & {FP-C} & {HL} & {LR} & {AR} & {RR} & {CSR} & {TR} \\
\midrule

\multirow[c]{16}{*}[-2ex]{LongVA-7B}
& \textsc{Base} & --
& 1.27 & 1.55 & 1.24 & 1.06 & 1.42 & 0.66 & 1.61 & 1.30 & 1.22 & 1.05 \\
\cmidrule(lr){2-13}

& \textsc{LoRA} & --
& 1.24 & 1.58 & 1.18 & 1.11 & 1.50 & 0.61 & 1.59 & 1.22 & 1.21 & 1.04 \\
\cmidrule(lr){2-13}

& \textsc{FFT} & --
& 1.15 & 1.43 & 1.10 & 0.99 & 1.45 & 0.58 & 1.35 & 1.14 & 1.01 & 0.98 \\
\cmidrule(lr){2-13}

& \textsc{Full-backbone} & --
& 1.25 & 1.56 & 1.23 & 1.13 & 1.56 & 0.63 & 1.42 & 1.32 & 1.20 & 1.08 \\
\cmidrule(lr){2-13}

& \textsc{All-attention} & --
& 1.28 & 1.56 & 1.27 & 1.14 & 1.26 & 0.60 & 1.65 & 1.30 & 1.20 & 1.08 \\
\cmidrule(lr){2-13}

& \multirow{5}{*}{\textsc{Random-k}} & 1
& 1.28 & 1.60 & 1.22 & 1.13 & 1.60 & 0.61 & 1.58 & 1.27 & 1.10 & 1.06 \\

& & 2
& 1.24 & 1.56 & 1.18 & 1.09 & 1.47 & 0.61 & 1.45 & 1.27 & 1.17 & 1.05 \\

& & 3
& 1.25 & 1.56 & 1.22 & 1.14 & 1.52 & 0.58 & 1.57 & 1.36 & 1.15 & 1.08 \\

& & 4
& 1.25 & 1.61 & 1.24 & 1.13 & 1.56 & 0.59 & 1.57 & 1.24 & 1.19 & 1.08 \\

& & 5
& 1.23 & 1.51 & 1.20 & 1.11 & 1.58 & 0.68 & 1.48 & 1.30 & 1.23 & 1.09 \\
\cmidrule(lr){2-13}

& \multirow{5}{*}{\shortstack{\textsc{Random-k}\\[-1pt]($\alpha$-tuned)}} & 1
& 1.28 & 1.60 & 1.22 & 1.13 & 1.60 & 0.61 & 1.58 & 1.27 & 1.10 & 1.06 \\

& & 2
& 1.25 & 1.59 & 1.18 & 1.12 & 1.58 & 0.55 & 1.51 & 1.29 & 1.14 & 1.07 \\

& & 3
& 1.27 & 1.53 & 1.20 & 1.13 & 1.56 & 0.54 & 1.51 & 1.33 & 1.21 & 1.09 \\

& & 4
& 1.23 & 1.59 & 1.24 & 1.15 & 1.56 & 0.63 & 1.61 & 1.32 & 1.17 & 1.07 \\

& & 5
& 1.24 & 1.58 & 1.22 & 1.08 & 1.58 & 0.60 & 1.58 & 1.28 & 1.14 & 1.07 \\
\cmidrule(lr){2-13}

& \ourcell  \ours{} & \ourcell  --
& \ourcell  1.29 & \ourcell  1.61 & \ourcell  1.25 & \ourcell  1.15 & \ourcell  1.40 & \ourcell  0.52 & \ourcell  1.57 & \ourcell  1.30 & \ourcell  1.16 & \ourcell  1.09 \\

\midrule

\multirow[c]{16}{*}[-2ex]{InternVL3-8B}
& \textsc{Base} & --
& 1.68 & 1.79 & 1.84 & 1.54 & 0.32 & 1.29 & 1.78 & 1.77 & 1.52 & 1.44 \\
\cmidrule(lr){2-13}

& \textsc{LoRA} & --
& 1.71 & 1.77 & 1.87 & 1.50 & 0.35 & 1.33 & 1.86 & 1.88 & 1.51 & 1.42 \\
\cmidrule(lr){2-13}

& \textsc{FFT} & --
& 1.69 & 1.82 & 1.88 & 1.51 & 0.37 & 1.41 & 1.86 & 1.87 & 1.65 & 1.41 \\
\cmidrule(lr){2-13}

& \textsc{Full-backbone} & --
& 1.71 & 1.74 & 1.85 & 1.46 & 0.37 & 1.32 & 1.85 & 1.81 & 1.54 & 1.44 \\
\cmidrule(lr){2-13}

& \textsc{All-attention} & --
& 1.71 & 1.81 & 1.88 & 1.54 & 0.32 & 1.32 & 1.78 & 1.86 & 1.58 & 1.44 \\
\cmidrule(lr){2-13}

& \multirow{5}{*}{\textsc{Random-k}} & 1
& 1.71 & 1.89 & 1.87 & 1.49 & 0.37 & 1.36 & 1.85 & 1.83 & 1.60 & 1.44 \\

& & 2
& 1.71 & 1.81 & 1.86 & 1.53 & 0.40 & 1.26 & 1.86 & 1.89 & 1.68 & 1.44 \\

& & 3
& 1.69 & 1.77 & 1.86 & 1.56 & 0.32 & 1.24 & 1.86 & 1.75 & 1.59 & 1.40 \\

& & 4
& 1.71 & 1.81 & 1.90 & 1.51 & 0.39 & 1.35 & 1.89 & 1.88 & 1.69 & 1.44 \\

& & 5
& 1.70 & 1.85 & 1.89 & 1.49 & 0.34 & 1.33 & 1.79 & 1.83 & 1.53 & 1.45 \\
\cmidrule(lr){2-13}

& \multirow{5}{*}{\shortstack{\textsc{Random-k}\\[-1pt]($\alpha$-tuned)}} & 1
& 1.71 & 1.77 & 1.89 & 1.53 & 0.32 & 1.33 & 1.87 & 1.79 & 1.70 & 1.46 \\

& & 2
& 1.70 & 1.85 & 1.89 & 1.56 & 0.32 & 1.31 & 1.85 & 1.88 & 1.54 & 1.43 \\

& & 3
& 1.71 & 1.84 & 1.89 & 1.52 & 0.32 & 1.22 & 1.85 & 1.80 & 1.60 & 1.45 \\

& & 4
& 1.70 & 1.79 & 1.87 & 1.56 & 0.37 & 1.37 & 1.83 & 1.82 & 1.59 & 1.42 \\

& & 5
& 1.69 & 1.85 & 1.86 & 1.52 & 0.37 & 1.31 & 1.88 & 1.85 & 1.65 & 1.45 \\
\cmidrule(lr){2-13}

& \ourcell  \ours{} & \ourcell  --
& \ourcell  1.72 & \ourcell  1.86 & \ourcell  1.86 & \ourcell  1.61 & \ourcell  0.37 & \ourcell  1.27 & \ourcell  1.85 & \ourcell  1.81 & \ourcell  1.73 & \ourcell  1.46 \\

\midrule

\multirow[c]{16}{*}[-2ex]{Qwen3-VL-4B}
& \textsc{Base} & --
& 1.50 & 1.76 & 1.59 & 1.22 & 0.66 & 1.47 & 1.76 & 1.42 & 1.48 & 1.08 \\
\cmidrule(lr){2-13}

& \textsc{LoRA} & --
& 1.56 & 1.87 & 1.63 & 1.28 & 0.92 & 1.50 & 1.95 & 1.79 & 1.41 & 1.24 \\
\cmidrule(lr){2-13}

& \textsc{FFT} & --
& 1.53 & 1.83 & 1.65 & 1.12 & 0.74 & 1.32 & 1.73 & 1.61 & 1.33 & 1.11 \\
\cmidrule(lr){2-13}

& \textsc{Full-backbone} & --
& 1.56 & 1.91 & 1.59 & 1.25 & 0.87 & 1.23 & 1.88 & 1.49 & 1.65 & 1.26 \\
\cmidrule(lr){2-13}

& \textsc{All-attention} & --
& 1.59 & 1.84 & 1.66 & 1.27 & 0.85 & 1.57 & 1.82 & 1.48 & 1.58 & 1.27 \\
\cmidrule(lr){2-13}

& \multirow{5}{*}{\textsc{Random-k}} & 1
& 1.59 & 1.91 & 1.66 & 1.23 & 0.94 & 1.55 & 1.86 & 1.44 & 1.64 & 1.28 \\

& & 2
& 1.56 & 1.85 & 1.59 & 1.29 & 0.87 & 1.33 & 1.81 & 1.66 & 1.43 & 1.30 \\

& & 3
& 1.41 & 1.80 & 1.46 & 1.08 & 0.68 & 1.26 & 1.59 & 1.36 & 1.35 & 1.08 \\

& & 4
& 1.59 & 1.84 & 1.67 & 1.27 & 0.76 & 1.42 & 1.80 & 1.65 & 1.78 & 1.36 \\

& & 5
& 1.62 & 1.91 & 1.67 & 1.28 & 1.00 & 1.35 & 1.80 & 1.58 & 1.64 & 1.34 \\
\cmidrule(lr){2-13}

& \multirow{5}{*}{\shortstack{\textsc{Random-k}\\[-1pt]($\alpha$-tuned)}} & 1
& 1.55 & 1.84 & 1.64 & 1.29 & 1.15 & 1.38 & 1.63 & 1.56 & 1.75 & 1.22 \\

& & 2
& 1.57 & 1.75 & 1.58 & 1.25 & 1.11 & 1.34 & 1.77 & 1.46 & 1.38 & 1.35 \\

& & 3
& 1.72 & 1.84 & 1.58 & 1.17 & 0.89 & 1.60 & 1.94 & 1.65 & 1.52 & 1.34 \\

& & 4
& 1.49 & 1.88 & 1.59 & 1.35 & 1.05 & 1.32 & 1.99 & 1.32 & 1.70 & 1.16 \\

& & 5
& 1.44 & 1.83 & 1.54 & 1.27 & 0.82 & 1.39 & 1.90 & 1.68 & 1.30 & 1.24 \\
\cmidrule(lr){2-13}

& \ourcell  \ours{} & \ourcell  --
& \ourcell  1.66 & \ourcell  1.94 & \ourcell  1.72 & \ourcell  1.29 & \ourcell  0.97 & \ourcell  1.63 & \ourcell  1.88 & \ourcell  1.58 & \ourcell  1.65 & \ourcell  1.38 \\

\bottomrule
\end{tabular}
\end{adjustbox}

\label{tab:mmbench_video}
\end{table*}

\begin{table*}[t]
\caption{\textbf{Full results (Accuracy \%) on Video-Holmes.}
CTI: Core Theme Inference;
IMC: Intention \& Motive Chaining;
MHR: Multimodal Hint Reasoning;
PAR: Physical Anomaly Reasoning;
SR: Social Reasoning;
TA: Timeline Analysis;
TCI: Temporal Causal Inference.}
\vspace{0.3em}
\centering
\small
\renewcommand{\arraystretch}{1.08}
\setlength{\tabcolsep}{4pt}

\begin{adjustbox}{max width=0.85\textwidth}
\begin{tabular}{c | l | c |
                S[table-format=2.1] |
                S[table-format=2.1]
                S[table-format=2.1]
                S[table-format=2.1]
                S[table-format=2.1]
                S[table-format=2.1]
                S[table-format=2.1]
                S[table-format=2.1]}
\toprule
Model & Variant & Run & {Overall} & {CTI} & {IMC} & {MHR} & {PAR} & {SR} & {TA} & {TCI} \\
\midrule

\multirow[c]{16}{*}[-2ex]{LongVA-7B}
& \textsc{Base} & --
& 15.2 & 9.3 & 18.8 & 18.1 & 11.3 & 24.3 & 3.0 & 15.8 \\
\cmidrule(lr){2-11}

& \textsc{LoRA} & --
& 14.5 & 9.6 & 17.0 & 19.0 & 10.3 & 24.7 & 1.5 & 13.2 \\
\cmidrule(lr){2-11}

& \textsc{FFT} & --
& 4.7 & 2.6 & 8.0 & 5.1 & 2.6 & 7.9 & 0.5 & 4.4 \\
\cmidrule(lr){2-11}

& \textsc{Full-backbone} & --
& 14.6 & 13.0 & 18.1 & 16.0 & 13.9 & 22.9 & 1.0 & 12.8 \\
\cmidrule(lr){2-11}

& \textsc{All-attention} & --
& 15.4 & 12.2 & 17.4 & 18.7 & 15.5 & 23.3 & 2.0 & 13.9 \\
\cmidrule(lr){2-11}

& \multirow{5}{*}{\textsc{Random-k}} & 1
& 16.7 & 12.2 & 22.5 & 17.5 & 13.4 & 26.7 & 1.0 & 17.6 \\

& & 2
& 10.0 & 7.0 & 12.3 & 11.4 & 7.7 & 18.2 & 1.5 & 8.1 \\

& & 3
& 13.7 & 13.0 & 13.4 & 14.8 & 11.3 & 26.4 & 2.5 & 9.5 \\

& & 4
& 19.8 & 16.3 & 25.4 & 24.4 & 12.9 & 20.5 & 12.0 & 21.6 \\

& & 5
& 12.3 & 8.9 & 13.4 & 13.0 & 12.4 & 19.5 & 5.5 & 11.0 \\
\cmidrule(lr){2-11}

& \multirow{5}{*}{\shortstack{\textsc{Random-k}\\[-1pt]($\alpha$-tuned)}} & 1
& 16.7 & 12.2 & 22.5 & 17.5 & 13.4 & 26.7 & 1.0 & 17.6 \\

& & 2
& 13.7 & 10.7 & 14.5 & 12.0 & 14.4 & 27.7 & 2.5 & 10.6 \\

& & 3
& 15.2 & 13.7 & 17.4 & 15.7 & 12.9 & 26.0 & 2.5 & 13.6 \\

& & 4
& 19.8 & 17.0 & 22.5 & 19.9 & 17.0 & 33.6 & 3.5 & 18.7 \\

& & 5
& 13.7 & 13.0 & 13.8 & 15.4 & 12.4 & 22.9 & 4.5 & 10.3 \\
\cmidrule(lr){2-11}

& \ourcell  \ours{} & \ourcell  --
& \ourcell  18.9 & \ourcell  17.0 & \ourcell  23.6 & \ourcell  20.5 & \ourcell  17.0 & \ourcell  29.8 & \ourcell  2.0 & \ourcell  16.1 \\

\midrule

\multirow[c]{16}{*}[-2ex]{InternVL3-8B}
& \textsc{Base} & --
& 36.5 & 34.8 & 42.4 & 34.0 & 35.6 & 44.5 & 31.0 & 31.5 \\
\cmidrule(lr){2-11}

& \textsc{LoRA} & --
& 36.2 & 36.3 & 36.2 & 31.3 & 34.0 & 42.5 & 37.5 & 35.9 \\
\cmidrule(lr){2-11}

& \textsc{FFT} & --
& 35.7 & 34.4 & 37.3 & 32.5 & 32.0 & 44.5 & 36.0 & 32.2 \\
\cmidrule(lr){2-11}

& \textsc{Full-backbone} & --
& 36.3 & 33.7 & 39.5 & 35.8 & 29.4 & 48.6 & 28.5 & 33.7 \\
\cmidrule(lr){2-11}

& \textsc{All-attention} & --
& 36.2 & 33.0 & 39.5 & 33.4 & 35.1 & 45.2 & 31.5 & 34.1 \\
\cmidrule(lr){2-11}

& \multirow{5}{*}{\textsc{Random-k}} & 1
& 37.4 & 35.9 & 40.6 & 36.1 & 32.5 & 51.0 & 26.5 & 34.1 \\

& & 2
& 36.1 & 36.7 & 39.9 & 32.2 & 33.5 & 48.3 & 29.0 & 30.4 \\

& & 3
& 35.1 & 32.2 & 36.2 & 34.0 & 32.0 & 47.3 & 31.5 & 30.0 \\

& & 4
& 37.4 & 37.0 & 38.8 & 34.0 & 33.0 & 50.3 & 32.0 & 33.7 \\

& & 5
& 37.4 & 34.4 & 40.6 & 34.6 & 36.6 & 51.0 & 33.5 & 29.3 \\
\cmidrule(lr){2-11}

& \multirow{5}{*}{\shortstack{\textsc{Random-k}\\[-1pt]($\alpha$-tuned)}} & 1
& 35.9 & 34.4 & 36.6 & 37.3 & 31.4 & 49.3 & 28.0 & 29.7 \\

& & 2
& 37.1 & 35.9 & 43.1 & 34.0 & 31.4 & 52.4 & 25.5 & 31.9 \\

& & 3
& 36.1 & 36.3 & 38.4 & 32.8 & 32.5 & 49.0 & 23.5 & 35.9 \\

& & 4
& 37.0 & 36.3 & 38.0 & 34.6 & 32.0 & 49.7 & 31.0 & 33.7 \\

& & 5
& 36.4 & 36.3 & 38.4 & 34.0 & 33.0 & 51.0 & 26.5 & 31.5 \\
\cmidrule(lr){2-11}

& \ourcell  \ours{} & \ourcell  --
& \ourcell  37.2 & \ourcell  34.1 & \ourcell  39.9 & \ourcell  37.0 & \ourcell  37.1 & \ourcell  49.7 & \ourcell  30.5 & \ourcell  29.3 \\

\midrule

\multirow[c]{16}{*}[-2ex]{Qwen3-VL-4B}
& \textsc{Base} & --
& 31.5 & 33.7 & 35.1 & 31.0 & 19.6 & 38.7 & 35.5 & 24.2 \\
\cmidrule(lr){2-11}

& \textsc{LoRA} & --
& 32.0 & 29.3 & 35.1 & 28.9 & 25.8 & 38.0 & 43.5 & 24.5 \\
\cmidrule(lr){2-11}

& \textsc{FFT} & --
& 30.7 & 28.1 & 40.2 & 28.3 & 24.7 & 33.6 & 39.0 & 21.6 \\
\cmidrule(lr){2-11}

& \textsc{Full-backbone} & --
& 31.2 & 33.0 & 36.2 & 29.5 & 25.8 & 41.8 & 27.5 & 21.6 \\
\cmidrule(lr){2-11}

& \textsc{All-attention} & --
& 31.9 & 29.6 & 34.1 & 29.8 & 21.1 & 45.9 & 37.0 & 23.4 \\
\cmidrule(lr){2-11}

& \multirow{5}{*}{\textsc{Random-k}} & 1
& 30.3 & 28.5 & 33.3 & 31.3 & 21.6 & 44.9 & 31.0 & 17.9 \\

& & 2
& 31.3 & 33.0 & 36.2 & 29.5 & 25.8 & 42.5 & 27.5 & 21.6 \\

& & 3
& 30.2 & 27.0 & 34.1 & 27.7 & 21.6 & 43.8 & 36.0 & 19.8 \\

& & 4
& 32.0 & 31.1 & 34.8 & 29.8 & 27.8 & 46.9 & 29.0 & 21.6 \\

& & 5
& 31.3 & 30.4 & 35.9 & 30.7 & 26.3 & 40.8 & 36.5 & 17.9 \\
\cmidrule(lr){2-11}

& \multirow{5}{*}{\shortstack{\textsc{Random-k}\\[-1pt]($\alpha$-tuned)}} & 1
& 32.1 & 29.6 & 33.7 & 30.4 & 24.2 & 45.5 & 35.5 & 23.8 \\

& & 2
& 30.8 & 28.9 & 32.2 & 28.6 & 22.2 & 43.8 & 34.0 & 23.8 \\

& & 3
& 32.6 & 31.9 & 35.5 & 30.4 & 23.7 & 46.2 & 37.0 & 21.6 \\

& & 4
& 31.7 & 30.0 & 34.4 & 29.8 & 24.7 & 43.8 & 34.5 & 22.7 \\

& & 5
& 31.2 & 29.6 & 35.1 & 30.1 & 25.3 & 41.1 & 36.0 & 20.5 \\
\cmidrule(lr){2-11}

& \ourcell  \ours{} & \ourcell  --
& \ourcell  32.9 & \ourcell  32.6 & \ourcell  38.4 & \ourcell  31.9 & \ourcell  23.2 & \ourcell  45.2 & \ourcell  32.0 & \ourcell  23.1 \\

\bottomrule
\end{tabular}
\end{adjustbox}
\label{tab:video_holmes_full}
\end{table*}

\end{document}